\documentclass[10pt,journal,compsoc]{IEEEtran}

\ifCLASSOPTIONcompsoc

  \usepackage[nocompress]{cite}
\else

  \usepackage{cite}
\fi

\ifCLASSINFOpdf
\else
\fi

\usepackage{amsmath}
\usepackage{booktabs} 
\usepackage{graphicx}
\usepackage{color}
\usepackage{float}
\usepackage{array}
\usepackage{multirow}
\usepackage{amssymb}
\usepackage{graphbox}

\usepackage{times}
\usepackage{epsfig}
\usepackage{graphicx}
\usepackage{amsmath}
\usepackage{amssymb}
\usepackage{lipsum}
\usepackage[bookmarks=false]{hyperref}
\hypersetup{
    colorlinks = true,
    citecolor  = blue,
    linkcolor  = blue,
    urlcolor   = blue,
}

\usepackage{url}           
\usepackage{booktabs}       
\usepackage{amsfonts}      
\usepackage{comment}
\usepackage{color}
\usepackage{multirow}
\usepackage{array}
\usepackage{graphbox}
\usepackage{booktabs} 
\usepackage{wrapfig}
\usepackage{enumitem}
\usepackage{bbding}
\usepackage{caption}

\newcommand{\etal}{\textit{et al.}}
\newcommand{\ie}{\textup{i.e.}}
\newcommand{\eg}{\textup{e.g.}}
\newcommand{\boldnum}[1]{\uppercase\expandafter{\romannumeral#1}}
\newcommand{\pamiparagraph}[1]{\vspace{0.2cm}\noindent\textbf{#1}\hspace{0.1cm} }
\renewcommand{\vec}[1]{\boldsymbol{#1}}
\newcommand{\bignum}[1]{\uppercase\expandafter{\romannumeral#1}}

\begin{document}

\title{PFENet++: Boosting Few-shot Semantic Segmentation with the Noise-filtered Context-aware Prior Mask}

\author{Xiaoliu Luo$^*$,
        Zhuotao Tian$^*$,
        Taiping Zhang$^\dagger$,
        Bei Yu,\\
        Yuan Yan Tang,~\IEEEmembership{Life~Fellow,~IEEE},
        Jiaya Jia,~\IEEEmembership{Fellow,~IEEE}
       
\IEEEcompsocitemizethanks{
\IEEEcompsocthanksitem  X.~Luo is with Chongqing University of Technology.
\IEEEcompsocthanksitem X.~Luo and T.~Zhang are with Chongqing University.
\IEEEcompsocthanksitem Z.~Tian, B.~Yu and J.~Jia are with The Chinese University of Hong Kong. 
\IEEEcompsocthanksitem Z.~Tian, B.~Yu and J.~Jia are with SmartMore.
\IEEEcompsocthanksitem Y.~Tang is with University of Macau.
\IEEEcompsocthanksitem $\dagger$ Corresponding author.
\IEEEcompsocthanksitem *Co-first authors with equal contributions, listed in alphabetical order.
}
}

\IEEEtitleabstractindextext{
\begin{abstract}
       In this work, we revisit the prior mask guidance proposed in ``Prior Guided Feature Enrichment Network for Few-Shot Segmentation''. The prior mask serves as an indicator that highlights the region of interests of unseen categories, and it is effective in achieving better performance on different frameworks of recent studies. However, the current method directly takes the maximum element-to-element correspondence between the query and support features to indicate the probability of belonging to the target class, thus the broader contextual information is seldom exploited during the prior mask generation. To address this issue, first, we propose the Context-aware Prior Mask (CAPM) that leverages additional nearby semantic cues for better locating the objects in query images. Second, since the maximum correlation value is vulnerable to noisy features, we take one step further by incorporating a lightweight Noise Suppression Module (NSM) to screen out the unnecessary responses, yielding high-quality masks for providing the prior knowledge. Both two contributions are experimentally shown to have substantial practical merit, and the new model named PFENet++ significantly outperforms the baseline PFENet as well as all other competitors on three challenging benchmarks PASCAL-5$^i$, COCO-20$^i$ and FSS-1000. The new state-of-the-art performance is achieved without compromising the efficiency, manifesting the potential for being a new strong baseline in few-shot semantic segmentation. 
Our code will be available at  \href{https://github.com/luoxiaoliu/PFENet2Plus}{Github}.
\end{abstract}

}

\maketitle

\IEEEdisplaynontitleabstractindextext

\IEEEpeerreviewmaketitle

\IEEEraisesectionheading{\section{Introduction}\label{sec:introduction}}
Deep learning has significantly boosted the performance of semantic segmentation. However, strong semantic segmentation models~\cite{deeplabv3,pspnet} heavily rely on the training with sufficient fully-labeled data, and they are hard to deal with new applications where novel classes are not witnessed during the training phase.

Few-shot segmentation (FSS)~\cite{shaban} aims at quickly adapting models to segment previously unseen categories on the query set with only a few annotations available in the support set. Models are episodically trained on base classes with sufficient annotations and then tested on novel classes. During testing, the novel categorical information is provided by the support set where only a few annotated samples are available, and models are required to locate the target objects in the query set based on the information given by the support set.

\begin{figure}
\centering
    \begin{minipage}   {0.97\linewidth}
        \centering
        \includegraphics [width=1\linewidth] 
        {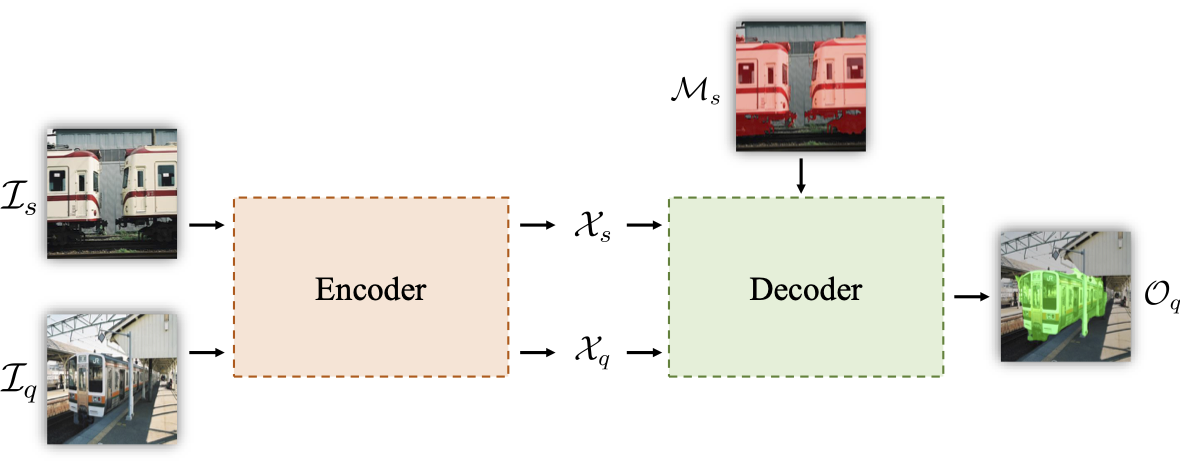}
    \end{minipage}
    \vspace{-0.2cm}    
    \caption{Abstract of recent FSS frameworks. $\mathcal{I}_s$ and $\mathcal{I}_q$ are the input support and query images, and $\mathcal{M}_s$ is the support mask. $\mathcal{X}_s$ and $\mathcal{X}_q$ denote the feature maps extracted by the encoder from the support and query images respectively. Then, the features are sent to the decoder, along with $\mathcal{M}_s$, to yield the prediction $\mathcal{O}_q$ on the query image $\mathcal{I}_q$, based on the categorical information provided by $\mathcal{M}_s$. The evaluation is conducted on the set of categories that are not witnessed during the training phase. }
    \vspace{-0.3cm}  
    \label{fig:encoder_decoder}
\end{figure}

Current FSS methods~\cite{panet,shaban,canet,ppnet,protomix} can be abstracted by a generic  encoder-decoder structure as demonstrated in Figure~\ref{fig:encoder_decoder}. The encoder is usually a deep convolutional network (\ie, VGG~\cite{vgg} and ResNet~\cite{resnet}) that processes the query and support images respectively to yield deep features. Then, the query and support features are sent to the decoder together with the support masks, and the decoder parses the input features and then outputs the predictions locating the target regions on the query samples. 

The recently proposed PFENet~\cite{pfenet} achieves promising performance on popular FSS benchmarks and it has served as strong baselines for the latest work~\cite{scl,asgnet,repri}. As observed in~\cite{canet,pfenet}, directly feeding the high-level features (\eg, \textit{layer 4} of ResNet) extracted from a fixed encoder to the trainable decoder results in performance deduction since the decoder will overly rely on the high-level features to make predictions during training, causing severe over-fitting to base classes. Therefore, PFENet instead transforms the high-level semantic cues into the class-agnostic prior mask by calculating the maximum one-to-one correlation responses between the query and support high-level features. 
Then, the prior mask is passed to the decoder along with the query and support middle-level features (\eg, \textit{layers 2-3} of ResNet), providing additional cues for better identifying the targets. The prior mask generation method is simple yet effective for achieving better results in FSS, while we find that two major bottleneck limits the performance.

\pamiparagraph{Underused high-level contextual cues.} Powerful scene parsing models achieve breakthrough improvements by adequately exploiting the semantic context from the high-level features (\eg, PPM~\cite{pspnet} and ASPP~\cite{aspp}). However, the current mask generation method of PFENet merely calculates the element-to-element correlations without considering the essential surrounding contextual information that is conducive to dense visual perception. Therefore, we alternatively use the Context-aware Prior Mask (CAPM) that is obtained by modeling the regional correlations. Unlike the element-wise correlations, feature patches encode more regional spatial information that could be used as additional hints for facilitating the dense labeling tasks.

\pamiparagraph{Noisy and unnecessary responses.}
The high-level prior guidance on the query sample is obtained by taking the maximum responses among all support features in~\cite{pfenet}, while it is observed that the maximum value is easily affected by noisy features that share local similarities (\ie, color and texture) but are with distinct semantic labels. Importantly, when more contextual cues are involved, the current parameter-free mask generation paradigm yields abundant unnecessary responses, making the generated masks unable to clearly indicate the region of interest. To alleviate this issue, we incorporate an effective yet efficient module named Noise Suppression Module (NSM) to screen out useless activations according to the correlation distribution between the query and support features, further improving the quality of the prior mask.

To this end, we combine PFENet with the proposed Context-aware Prior Mask (CAPM) and Noise Suppression Module (NSM), and the enhanced model named PFENet++ significantly outperforms PFENet as well as all other competitors in both 1- and 5-shot settings. Though only a few additional learnable parameters are introduced, the proposed CAPM and NSM still well generalize to the new categories that even do not exist in the ImageNet~\cite{imagenet} used for pre-training the feature extractor.
Our contributions in this paper are summarized as follows.

\begin{itemize}
    \item To our best knowledge, the proposed Context-aware Prior Mask (CAPM) is the first design that exploits the regional contextual correlation between the query and support features to address the few-shot segmentation problem.
    \item By mitigating redundant irrelevant correlation responses, the Noise Suppression Module (NSM) further boosts the improvement brought by CAPM.
    \item PFENet++ reaches a new state-of-the-art performance on popular benchmarks without deprecating the model efficiency, and the new designs also bring considerable improvements to other latest methods.
\end{itemize}

\section{Related Work}
\vspace{0.2cm}\noindent\textbf{Semantic Segmentation} \\
Semantic segmentation is a fundamental yet challenging task that requires accurate dense labeling. Recent architectures~\cite{deconvnet,segnet,unet,hrnet_cvpr,hrnet_pami,tian2023cac,lai2023lisa} are originated from FCN~\cite{fcn} that is the first framework designed for semantic segmentation. Contextual information helps identify individual elements based on the surrounding hints, thus the receptive field is essential for semantic segmentation. To this end, the dilated convolution~\cite{deeplab,dilation,aspp,denseaspp}, global pooling~\cite{parsenet} and pyramid pooling~\cite{pspnet,icnet,strippool} are adopted to help enlarge the receptive field and they have achieved considerable improvement. Meanwhile, to effectively leverage long-distance semantic relations, attention-based models~\cite{encnet,danet,ocr,ccnet,psanet} have thrived, reaching a new state-of-the-art performance. However, powerful segmentation models cannot well address the previously unseen categories without updating the model parameters. 

\vspace{0.2cm}\noindent\textbf{Few-Shot Learning} \\
Few-shot learning methods perform classification on the novel categories while only a small amount of labeled data is provided. Recent solutions are mainly based on meta-learning \cite{memory_match,maml,leo} that aims to get a model that can quickly generalize to new downstream applications within a few adaption steps, and metric-learning \cite{matchingnet,relationnet,prototype_cls,deepemd,imprint_cvpr18,dynamic_noforget} that learns to yield discriminative representations for novel categories.
Moreover, considering the data-driven literature of deep learning, data augmentation techniques help models achieve better performance by synthesizing new samples or features based on the few labeled data \cite{hallucinate_saliency,hallucinating,imaginary}. 
Though few-shot learning has made tremendous progress on image recognition, without considering the contextual issues, directly applying the representative few-shot learning methods (\ie, Prototypical Network~\cite{prototype_cls} and Relation Network~\cite{relationnet}) to address few-shot segmentation achieve less satisfying results as verified by the baselines of PL~\cite{prototype_seg} and CANet~\cite{canet}. 

\vspace{0.2cm}\noindent\textbf{Few-Shot Segmentation} \\
Few-Shot Segmentation (FSS) requires model to quickly segment the target region with only a few annotations~\cite{guide,selftuning,differentiable,fss1000,fewshot3d}.
OSLSM \cite{shaban} formally introduces this setting in segmentation and provides a solution by imprinting the classifier's weights for each task. The idea of Prototypical Network~\cite{prototype_cls} is used in PL \cite{prototype_seg} whose predictions are made based on the cosine similarity between the query pixels and support prototypes. Additionally, prototype alignment regularization is introduced in PANet \cite{panet} to help rectify the prediction. The prototype mixture models (PMMs)~\cite{protomix} correlate diverse image regions with multiple prototypes to enforce the prototype-based semantic representation with the Expectation-Maximization (EM) algorithm.
Predictions can be also generated by convolutions, analogous to the relation module proposed in~\cite{refinenet}. CANet~\cite{canet} concatenates the support and query features and adopts the Iterative Optimization Module (IOM) to accomplish the prediction refinement. PPNet~\cite{ppnet} constructs partial support prototypes based on super-pixels.  PFENet \cite{pfenet} exploits the prior mask guidance obtained from the pre-trained backbone to help locate the target region, and a Feature Enrichment Module (FEM) to tackle the spatial inconsistency between the query and support samples, respectively. RePRI~\cite{repri} proposes a transductive inference strategy that better leverages the support-set supervision than other existing methods.

More recently,~\cite{lu2021simpler,wu2021learning,CMN,hsnet,kang2022ifsl,lang2022bam,dpcn,ntre,gfsseg,fan2022ssp,hong2022cost,dgpnet,HMNet,shi2022dense,AAT,ddac,peng2023hierarchical,rcnet,MIANet,Hajimiri2022ASB,LOP,kang2023distilling} further boost the performance. Specifically, HSNet~\cite{hsnet} employs multi-level feature correlation and efficient 4D convolutions to extract a range of features from different levels of intermediate convolutional layers, resulting in the generation of a collection of 4D correlation tensors. BAM~\cite{lang2022bam} introduces an additional branch (base learner) alongside the conventional FSS model (meta learner) with the explicit purpose of identifying the targets of base classes. The final segmentation prediction is obtained by adaptively integrating the coarse results produced by these two learners, allowing for a more accurate segmentation. GFS-Seg~\cite{gfsseg} analyzes the generalization ability of segmentation models to simultaneously recognize novel categories with very few examples as well as base categories with sufficient examples. VAT~\cite{hong2022cost} proposes a 4D Convolutional Swin Transformer to address the problem arose from tokenization of a correlation map from transformer processing.
Our contributions in this paper mainly follow the principles of the prior mask guidance proposed in PFENet~\cite{pfenet}.

\section{Preliminaries}
Before formally introducing our method, we start with the task definition of Few-shot Segmentation (FSS) in Sec.~\ref{sec:task_description}, followed by the introduction of the prior mask guidance proposed in PFENet~\cite{pfenet} in Sec.~\ref{sec:prior_mask}. 

\subsection{Task Description}
\label{sec:task_description}
The few-shot semantic segmentation task is associated with two sets, i.e., the query set $\mathcal{Q}$ and support set $\mathcal{S}$. Given $\mathcal{K}$ samples from support set $\mathcal{S}$, the goal is to segment the area of unseen class $\mathcal{C}_{test}$ from each query image $\mathcal{I}_q$ in the query set. 

Models are trained on base classes $\mathcal{C}_{train}$  and tested on previously unseen (novel) classes $\mathcal{C}_{test}$ in episodes $(\mathcal{C}_{train} \cap~  \mathcal{C}_{test} = \varnothing)$. The episode paradigm for FSS is proposed in  \cite{shaban} for training and evaluation. Each episode is formed by a support set and a query set of class $\mathcal{C}_i$. The support set $\mathcal{S}$ has $\mathcal{K}$ samples $\mathcal{S}=\{\mathcal{S}_1, \mathcal{S}_2, ..., \mathcal{S}_{\mathcal{K}}\}$, named `$\mathcal{K}$-shot scenario'. Each support sample $\mathcal{S}_i$ is a pair of $\{\mathcal{I}_s, \mathcal{M}_s\}$ where $\mathcal{I}_s$ and $\mathcal{M}_s$ are the support image and pixel-wise annotation of $\mathcal{C}_i$ respectively. For the query set, $\mathcal{Q}=\{\mathcal{I}_q, \mathcal{M}_q\}$ where $\mathcal{I}_q$ is the query image and $\mathcal{M}_q$ is the ground truth indicating the target belonging to class $\mathcal{C}_i$. The query-support pair $\{\mathcal{I}_q, \mathcal{S}\} = \{\mathcal{I}_q, \mathcal{I}_s, \mathcal{M}_s\}$ forms the input batch.

\subsection{Revisit the Prior Mask Guidance}
\label{sec:prior_mask}
In CANet~\cite{canet}, Zhang~\etal~empirically observe that, within the encoder-decoder structure, directly sending the middle-level features extracted from a fixed encoder (\eg, ResNet~\cite{resnet} and VGG~\cite{vgg}) to the decoder performs much better than the high-level counterparts on few-shot segmentation, since the middle-level ones constitute object parts shared by unseen classes. 

Tian~\etal~\cite{pfenet} give an alternative explanation that the semantic information contained in high-level features is more class-specific than the middle-level features, and therefore the decoder parameters are prone to overly rely on high-level features of the base classes to optimize the training objectives better, causing inferior generalization ability on unseen classes. However, contradicting these findings in FSS, advanced generic semantic segmentation frameworks~\cite{deeplabv3+,pspnet,ocr} are instead designed for better exploiting the high-level semantic cues to have superior performance. Therefore, for the purpose of adequately making use of the high-level hints that could cause severe over-fitting issues, PFENet~\cite{pfenet} transforms the ImageNet~\cite{imagenet} pre-trained high-level features into a class-agnostic prior mask that merely indicates the probability of pixels belonging to the target category.

Concretely, let ${\mathcal{F}}$ denote the feature extractor, and ${\mathcal{I}_q}$ and ${\mathcal{I}_s}$ are query and support images respectively. We can obtain the $d$-dimensional high-level features as:
\begin{equation}
\label{eqn:original_priormask_1}
\mathcal{X}_q = \mathcal{F}(\mathcal{I}_q), \quad \mathcal{X}_s = \mathcal{F}(\mathcal{I}_s)\odot \mathcal{M}_s,
\end{equation}
where $\odot$ is the Hadamard product. Let $h$ and $w$ represent the height and width of the feature map. The irrelevant area on the reshaped support high-level feature $\mathcal{X}_s$ $\in \mathbb{R}^{hw\times d}$ is set to zero by the reshaped binary support mask $\mathcal{M}_s$ $\in \mathbb{R}^{hw\times 1}$, to make sure that the prior mask of $\mathcal{I}_q$ only correlates with the target region of $\mathcal{I}_s$.
The element-to-element correlation matrix $\mathcal{R}$ $\in \mathbb{R}^{hw\times hw}$ is subsequently formed by calculating the cosine-similarity values between all query and support high-level features:
\begin{equation}
\label{eqn:similarity_matrix}
\mathcal{R} = \phi(\mathcal{X}_q) \phi(\mathcal{X}_s)^T,
\end{equation}
where $\phi$ represents the L2 normalization.
Then, the prior mask $\mathcal{Y}_q$ $\in \mathbb{R}^{hw\times 1}$ of the query image can be yielded by taking the maximum correlation values among all support features in Eq.~\eqref{eqn:argmax}, followed by the min-max normalization as Eq.~\eqref{eqn:norm} with $\epsilon=1e-7$.
\begin{equation}
\label{eqn:argmax}
\mathcal{Y}_q = \mathop{\max}_{j} \mathcal{R}(i,j), \quad i,j\in \{1, 2, ..., hw\},  
\end{equation}
\begin{equation}
\label{eqn:norm}
\mathcal{Y}_q = \frac{\mathcal{Y}_q - \min(\mathcal{Y}_q)}{\max(\mathcal{Y}_q) - \min(\mathcal{Y}_q) + \epsilon}.
\end{equation}
\begin{figure*}[!t]
\centering
    \begin{minipage}   {0.99\linewidth}
        \centering
        \includegraphics [width=1\linewidth] 
        {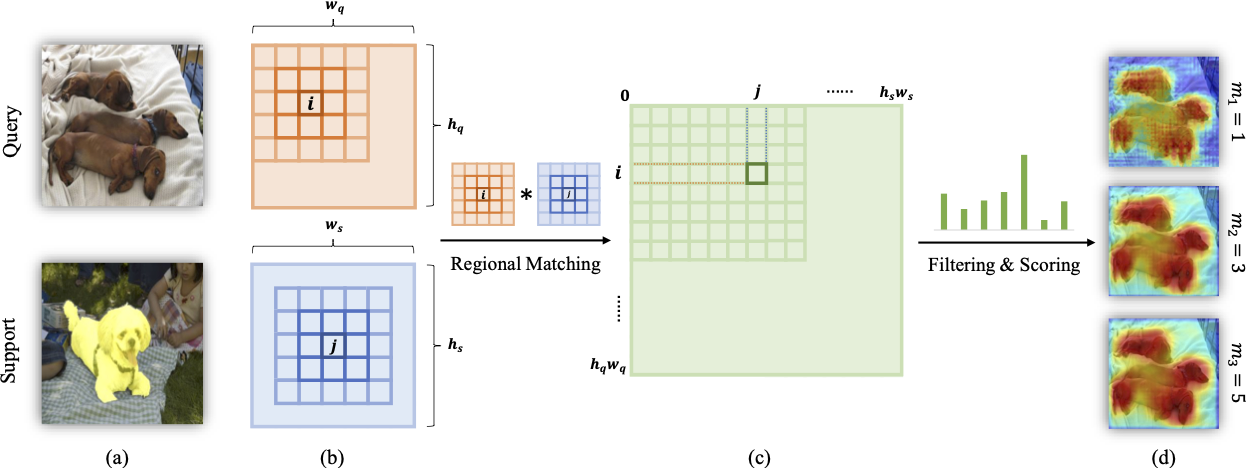}
    \end{minipage}
    \vspace{-0.2cm}
    \caption{Illustration of the new prior generation pipeline that is composed of two steps: 1) regional matching and 2) filtering \& scoring. (a) and (b) demonstrate the input query and support images and their features respectively. (c) is the correlation matrix, and (d) shows prior masks generated by different patch sizes. As introduced in Sec.~\ref{sec:capm}, the first step,  regional matching, aims at yielding the context-aware prior masks (CAPM) that incorporate broader information for better locating the objects belonging to unseen categories. The second step filtering \& scoring is accomplished by the proposed Noise Suppression Module (NSM) that rectifies the activation values of support samples so as to provide high-quality prior guidance, as described in Sec.~\ref{sec:nsm}.}
    \vspace{-0.2cm}
    \label{fig:method}
\end{figure*}

\section{Method}
In this following, the motivations and details of our proposed Context-aware Prior Mask (CAPM) and Noise Suppression Module (NSM) are shown in Sec.~\ref{sec:capm} and Sec.~\ref{sec:nsm} respectively. The overall pipeline is illustrated in Fig.~\ref{fig:method} where regional matching and noise suppression are performed to yield more informative prior masks.

\subsection{Context-aware Prior Mask}
\label{sec:capm}
The prior mask $\mathcal{Y}_q$ obtained by Eqs.~\eqref{eqn:argmax} and ~\eqref{eqn:norm} implies the similarities between the query features and their most relevant support features, thus the higher the values on $\mathcal{Y}_q$, the more likely the corresponding regions belong to the target. However, these correlation values on $\mathcal{Y}_q$ are obtained by one-to-one matching without considering broader contextual information. Locations belonging to the background of the query image might also have strong responses on $\mathcal{Y}_q$ as long as they are locally similar to only one foreground element of the support sample, making the prior mask less effective in indicating the region of interest. 
As shown by examples in Fig.~\ref{fig:visual_compare_prior} (a), the ones yielded by the one-to-one matching scheme cannot well handle these cases because the surrounding non-target regions on the query sample might also have high correlation values with the locally identical support features, making the normalization process less discriminative in highlighting the region of interest on the prior mask.
Therefore, the underutilized contextual information might be an inherent bottleneck limiting further improvements that could be brought by high-level features. 

To make the best of the nearby hints, one might directly apply context enrichment modules like PPM~\cite{pspnet} and ASPP~\cite{aspp} to the high-level query and support features respectively to help each element be more aware of the surroundings before being used for the prior mask generation. Despite the fact that these modules are widely found to be conducive to normal semantic segmentation frameworks, they introduce considerable trainable parameters that might cause over-fitting issues, which has been verified by experimental results in Table~\ref{tab:ablation_study}. 
For this reason, instead of the commonly used contextual enhancement modules that cause performance deduction, we adopt a regional matching strategy to generate the Context-aware Prior Mask (CAPM).

\pamiparagraph{Regional matching.}
The idea of CAPM is straightforward: it calculates the regional correlations rather than the original element-to-element ones. For generality, we assume that the query and support features might have different spatial sizes, denoted as $h_q\times w_q$ and $h_s\times w_s$.

Formally, $\mathcal{X}_q$ $\in \mathbb{R}^{h_q\times w_q\times d}$ and $\mathcal{X}_s$ $\in \mathbb{R}^{h_s\times w_s\times d}$ denote the $d$-dimensional query and support features obtained via Eq.~\eqref{eqn:original_priormask_1}, and we assume that they have been already L2-normalized by $\phi$. Then, regional matching can be accomplished by measuring the patch-wise similarities between $\mathcal{X}_s$ and $\mathcal{X}_q$, formulated as:
\begin{equation}
\label{eqn:patch_similarity}
\mathcal{R}^{c}(\vec{i}, \vec{j}) = \sum_{\vec{o} \in [-m,m]\times[-m,m]} \langle \mathcal{X}_q(\vec{i} + \vec{o}),\, \mathcal{X}_s(\vec{j} + \vec{o}) \rangle.
\end{equation}

In Eq.~\eqref{eqn:patch_similarity}, $m$ is the patch size and $\vec{o} \in [-m,m]\times[-m,m]$ represents the set of feasible offset tuples within the $m \times m$ patch window. The patch center positions for $\mathcal{X}_q$ and $\mathcal{X}_s$ are denoted by $\vec{i} \in [0,h_q]\times[0,w_q]$ and $\vec{j} \in [0,h_s]\times[0,w_s]$ respectively. $\mathcal{X}_q(\vec{i} + \vec{o})$ and $\mathcal{X}_s(\vec{j} + \vec{o})$ are $d$-dimensional vectors, thus $\mathcal{R}^{c}(\vec{i}, \vec{j}) \in [-1, 1]$ is scalar value representing the patch-wise similarity. If $\vec{i} + \vec{o} \notin [0,h_q]\times[0,w_q]$ or $\vec{j} + \vec{o} \notin [0,h_s]\times[0,w_s]$, zeros are padded.  
 The superscript $c$ in $\mathcal{R}^{c}$ denotes that the contextual information is better exploited by the proposed regional matching process.

In consequence, with Eq.~\eqref{eqn:patch_similarity}, we can obtain the regional similarity matrix $\mathcal{R}^{c} \in \mathbb{R}^{h_q\times w_q\times h_s \times w_s}$ that can be reshaped to $\mathcal{R}^{c} \in \mathbb{R}^{h_q w_q\times h_s w_s}$ and then processed by Eq.~\eqref{eqn:argmax} and Eq.~\eqref{eqn:norm} sequentially to produce the new context-aware prior mask $\mathcal{Y}^{c}_q \in \mathbb{R}^{h_q w_q\times 1}$ that exploits more contextual cues to help identify the target objects.

\pamiparagraph{Does a single large patch take all?}
\label{sec:option}
It is obvious that the larger the patch size is, the more contextual information could be exploited to yield the context-aware prior mask. Intuitively, $1\times 1$ and $3\times 3$ patches can be covered by a $5\times 5$ patch, thus a natural option is to apply the regional matching with a single large patch size $m$ that can reach an optimal trade-off between the performance and efficiency. 

Nevertheless, the greater $m$ not only brings additional computation overhead, it might also introduce redundant information that is detrimental for revealing the local details throughout the regional matching process, leading to sub-optimal performance. As verified by our experimental results in Table~\ref{tab:scale_capm}, the regional matching with $m$=3 outperforms $m$=1 thanks to the contextual awareness, while results of $m$=5 do not further advance the ones yielded by $m$=1 and $m$=3.
To this end, in order to utilize the context-aware prior mask without deprecating the local discrimination ability, we propose an alternative way that accomplishes the regional matching with multiple patch sizes so as to large patch captures nearby context and small patch mines finer details.

Let $M$=$\{m_1, m_2, ..., m_{|M|}\}$ be the set of patch sizes adopted for regional matching, and we can accordingly obtain $|M|$ prior masks $\{\mathcal{R}^{c}_1, \mathcal{R}^{c}_2, ..., \mathcal{R}^{c}_{|M|}\}$ via Eq.~\eqref{eqn:patch_similarity}. Then, the multi-patch method produces the context-aware prior mask $\mathcal{Y}^{c}_q \in \mathbb{R}^{h_q w_q\times |M|}$ with all matrices contained in the set $\{\mathcal{R}^{c}_1, \mathcal{R}^{c}_2, ..., \mathcal{R}^{c}_{|M|}\}$. This process can be formulated as:
\begin{equation}
\label{eqn:multi_patch_similarity}
\begin{aligned}
&\mathcal{R}^{c} = \mathcal{R}^{c}_1 \oplus ... \oplus \mathcal{R}^{c}_m \oplus ... \oplus \mathcal{R}^{c}_{|M|} \\
&\mathcal{Y}^{c}_{q, m} = \mathop{\max}_{j} \mathcal{R}^{c}(i,j,m) \\
&i,j\in \{1, 2, ..., hw\}, \quad m \in \{1,...,|M|\}, 
\end{aligned}
\end{equation}
where $\oplus$ denotes the concatenation operation between $|M|$ matrices $\mathcal{R}^{c}_m \in \mathbb{R}^{h_q w_q\times h_s w_s}$ to form $\mathcal{R}^{c} \in \mathbb{R}^{h_q w_q\times h_s w_s \times |M|}$, and the prior mask $\mathcal{Y}^{c}_q \in \mathbb{R}^{h_q w_q \times |M|}$ is yielded by taking the maximum values among the second dimension of $\mathcal{R}^{c}$.

\subsection{Noise Suppression Module}
\label{sec:nsm}

\pamiparagraph{Motivation.}
In~\cite{pfenet}, taking the maximum response is found to be conducive to revealing most of the potential target on a query image, because the maximum value indicates that the support image contains at least one pixel or region that has close semantic relation to the query pixel. However, these extreme values are vulnerable to the noisy support features that are locally identical to the query features but are actually from distinct classes. Moreover, in complex scenes, the noisy support features might cause abundant false alarms on the prior mask, making it fails to highlight the potential region on the query image. 

To alleviate the above issues, we need a module that helps screen out unnecessary responses. Fundamentally, the module design should be \textit{content-aware} such that it can dynamically mitigate the responses of irrelevant regions, and on the other hand, it should also be \textit{class-agnostic}, in an attempt to avoid over-fitting to the base classes during training. 
With these two essential design principles, the proposed Noise Suppression Module (NSM) brings considerable performance gain to the baseline models.

With the aim of being context-aware and class-agnostic, the proposed Noise Suppression Module (NSM) is applied to the query-support similarity matrices $\mathcal{R} \in \mathbb{R}^{h_q w_q\times h_s w_s}$ and $\mathcal{R}^{c} \in \mathbb{R}^{h_q w_q\times h_s w_s \times |M|}$, since there exists no class-sensitive cues contained in the similarity matrices but only the correlation information of the support sample that has been outlined by its ground-truth mask in the second dimension with a size of $h_s w_s$. Moreover, the correlation information is exactly the context provider that tells NSM which part of the support sample is mostly relevant to the majority features of the query sample from a holistic perspective, and thus NSM can simply suppress the rest suspected to cause noisy responses on the prior mask.

\begin{figure}[!t]
\centering
    \begin{minipage}   {0.97\linewidth}
        \centering
        \includegraphics [width=1\linewidth] 
        {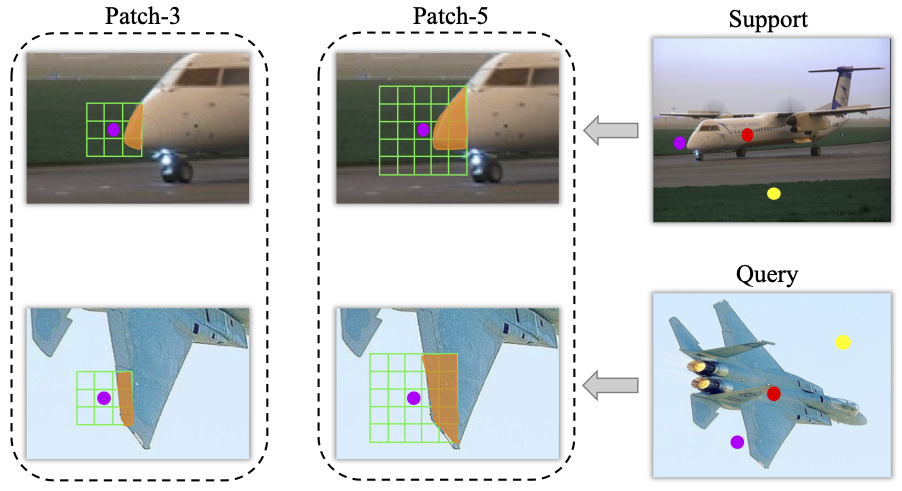}
    \end{minipage}
    \vspace{-0.2cm}    
    \caption{Illustration of the $3\times3$ and $5\times5$ region matching. Three types of pixels exist: background, edge-background and foreground pixels, denoted as yellow, purple and red circles respectively.}
    \label{fig:noise}
\end{figure}

\begin{figure}[!t]
\centering
    \begin{minipage}   {0.97\linewidth}
        \centering
        \includegraphics [width=1\linewidth] 
        {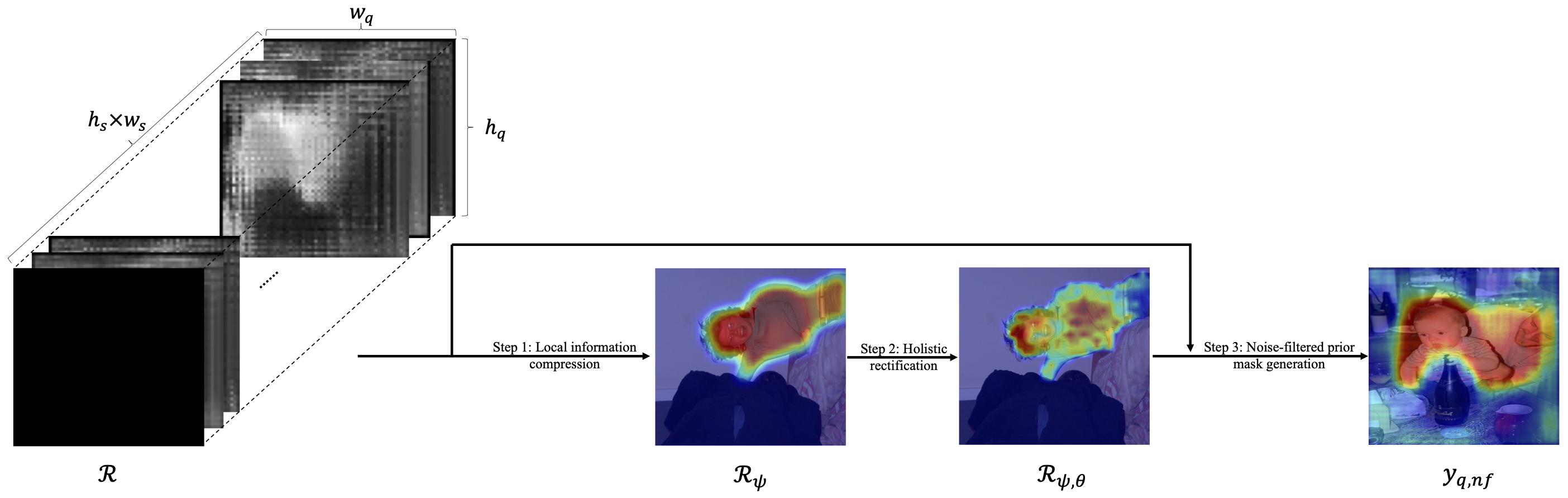}
    \end{minipage}
    \caption{The pipeline of Noise Suppression Module (NSM).}
    \label{fig:nsm}
\end{figure}

\pamiparagraph{Overview.}
The Noise Suppression Module (NSM) has three steps: 1) local information compression, 2) holistic rectification and 3) noise-filtered prior mask generation. As illustrated in Figure~\ref{fig:nsm}, the former two steps, from local and global views respectively, yield a rectifier $\mathcal{R}_{\psi,\theta}$ that adaptively adjusts the contributions of support features on the correlation matrices in the last step. To this end,  the adverse effects brought by the noisy activations are mitigated. Details are as follows.

\pamiparagraph{Step 1: Local information compression.}
In provision for estimating the importance of each support feature, an information concentrator $\Psi$ is used as a means to show how much the individual support feature is correlated to the query features. Specifically, we have
\begin{equation}
\label{eqn:concentrator}
\mathcal{R}_{\psi} = \Psi(\mathcal{R}),
\end{equation}
where the first dimension of $\mathcal{R} \in \mathbb{R}^{h_q w_q\times h_s w_s}$ has been compressed to 1 in $\mathcal{R}_{\psi} \in \mathbb{R}^{1\times h_s w_s}$. The concentrator $\Psi$ can be either a parameterized module or an operation that takes the average values. Besides,  if the regional matching with $|M|$ patch sizes is adopted, Eq.~\eqref{eqn:concentrator} can be also applied to $\mathcal{R}^{c} \in \mathbb{R}^{h_q w_q \times h_s w_s \times |M|}$ by processing $|M|$ matrices separately to yield the compressed output $\mathcal{R}^{c}_{\psi} \in \mathbb{R}^{1 \times h_s w_s \times |M|}$.

The concentrator $\Psi$ merely probes the correlations between all query features and individual support features, accordingly, it only examines the global context of the query features to foreground the essential individual support features. As the categorical information is provided by the support sample, the spatial information (\ie, correlation distribution) of the support sample should also be considered to facilitate sweeping away those ``bad'' ones that might cause undesired high responses on the prior mask. 

\pamiparagraph{Step 2: Holistic rectification.}
To leverage the contextual cues of support features in a class-agnostic way, we look into the importance of each support feature from a holistic perspective. Specifically, a rectification module $\Theta$ is proposed to adjust the importance of support features, and this process can be written as
\begin{equation}
\label{eqn:holistic_refine}
\mathcal{R}_{\psi,\theta} = \Theta(\mathcal{R}_{\psi}),
\end{equation}
where the module $\Theta$ in Eq.\eqref{eqn:holistic_refine} is composed of a few learnable light-weight fully-connected layers. $\Theta$ yields a rectifier $\mathcal{R}_{\psi,\theta} \in \mathbb{R}^{1 \times h_s w_s}$ that adjusts the contributions of each support feature on the similarity matrix $\mathcal{R} \in \mathbb{R}^{h_q w_q\times h_s w_s}$ in the next step.

\pamiparagraph{Step 3: Noise-filtered prior mask generation.}
The rectifier $\mathcal{R}_{\psi,\theta} \in \mathbb{R}^{1 \times h_s w_s}$ can be deemed as a set of values weighing the support elements, thus the noise-filtered prior mask $\mathcal{Y}_{q, nf}$ is obtained as
\begin{equation}
\label{eqn:noise_filter_mask_generation}
\mathcal{Y}_{q, nf} = \mathcal{R}\mathcal{R}_{\psi,\theta}^{T} ,
\end{equation}
by which the interfering responses on the original correlation matrix $\mathcal{R}$ caused by irrelevant support elements are suppressed by the rectifier $\mathcal{R}_{\psi,\theta}$. The visual comparison is shown in Fig.~\ref{fig:visual_compare_prior} (b) and (c). 

Similarly, $\mathcal{Y}_{q, nf}^{c} \in \mathbb{R}^{h_q w_q \times |M|}$ can be obtained by applying Eq.~\eqref{eqn:holistic_refine} and Eq.~\eqref{eqn:noise_filter_mask_generation} to $\mathcal{R}^{c} \in \mathbb{R}^{h_q w_q \times h_s w_s \times |M|}$ by independently processing $|M|$ matrices, if the multi-patch regional matching is adopted as introduced in Sec.~\ref{sec:capm}.

\subsection{Interpretations regarding CAPM and NSM. }
\pamiparagraph{Noisy responses caused by CAPM.}
As shown in Figure~\ref{fig:noise}, there are three sets of pixels in the support and query images: background, edge-background and foreground pixels, and they are denoted by yellow, purple, and red circles respectively. During the prior generation, all background pixels in the support image are simply ignored by the support mask, while the regional matching with $3\times3$ and $5\times 5$ patches will bring noises to the edge-background pixels by involving partial foreground pixels as shown in the orange regions in Figure~\ref{fig:noise}. 

\begin{figure}[!t]
\centering
    \begin{minipage}   {0.97\linewidth}
        \centering
        \includegraphics [width=1\linewidth] 
        {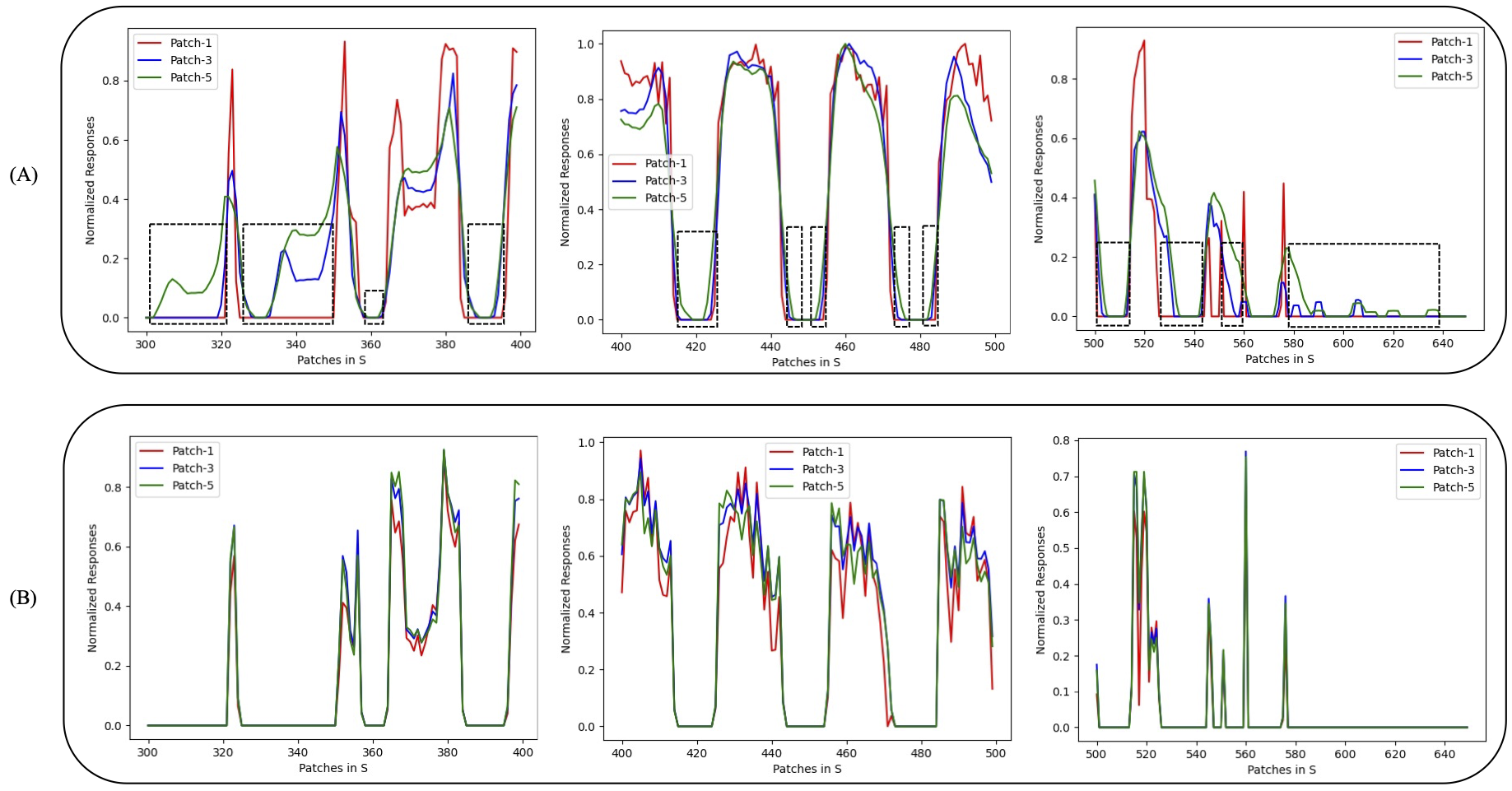}
    \end{minipage}
    \caption{Statistical comparison regarding the distributions of the inputs (A) and outputs (B) of the rectification module $\Theta$. Note that the responses are normalized for highlighting the relative changes. }
    \label{fig:statistic_compare}
\end{figure}

Specifically, for the query image, we take the patch of the edge-background pixel in the query image to match with all pixels in the support image where the background pixels have been masked to zero, while the edge-background pixel in the query image will have some responses on the prior mask due to the patch that includes partial foreground pixels. However, these responses are confusing because the query edge-background pixels belong to the background, making the prior mask difficult to highlight the foreground target by blurring the boundary regions.
As a result, without NSM, the blurred region expands as the patch size increases as demonstrated in Figure~\ref{fig:statistic_compare}.

\pamiparagraph{Understanding the intermediate results of NSM. }
The pipeline in Figure~\ref{fig:nsm} shows that, in Step 1 (Local information compression), the first dimension of the correlation matrix $\mathcal{R} \in \mathbb{R}^{h_q w_q\times h_s w_s}$ (only a single patch is considered for simplicity) is compressed to 1 in $\mathcal{R}_{\psi} \in \mathbb{R}^{1\times h_s w_s}$ by a concentrator $\Psi$ that takes the average values as $\mathcal{R}_{\psi} = \Psi(\mathcal{R})$. 
$\mathcal{R}_{\psi}$ stores the statistics that tell how much each part of the support sample is related to the elements of the query sample. 
Then, Step 2 rectifies the regional responses on the support sample by holistically analyze the distributions in ${\mathcal{R}}_{\Psi}$, such that informative regions of the support sample are assigned with greater importance by $\mathcal{R}_{\psi,\theta} = \Theta(\mathcal{R}_{\psi})$ ($\mathcal{R}_{\psi,\theta} \in \mathbb{R}^{1 \times h_s w_s}$). Finally, the activations are weighted by $\mathcal{R}_{\psi,\theta}$ to obtain the noise-filtered prior mask $\mathcal{Y}_{q, nf} = \mathcal{R}\mathcal{R}_{\psi,\theta}^{T}$ ($\mathcal{Y}_{q, nf} \in \mathbb{R}^{h_q w_q, 1}$). 

To investigate the effect of the proposed NSM, 
statistical examples are presented in Figure~\ref{fig:statistic_compare} where we use line charts to visualize the results $\mathcal{R}_{\psi} \in \mathbb{R}^{1\times h_s w_s}$ of the concentrator $\Psi$, and the red, blue and green colors denote $1\times 1$, $3\times 3$ and $5\times 5$ patches, respectively. The row represents the patch indexes in the support feature map with the size of $h_s\times w_s$, and the column represents the averaged similarity values between each patch in the support feature map and all patches in the query feature maps.

Concretely, for the patch size $1\times 1$, the averaged similarity between each background pixel in the support image and all pixels in the query image should be close to 0 since the background pixels in the support image has been filtered out by the support mask. However, when performing the $3\times3$ and $5\times5$ region matching, some averaged similarity values between the patches of the support edge-background patches and query patches are not 0 because the patches of edge-background pixels may include partial foreground pixels. As shown in the black dashed boxes of the top three examples in Figure~\ref{fig:statistic_compare} (A) , it can be observed that even when the red line (Patch-1) is close to 0, higher responses are observed on the blue (Patch-3) green (Patch-5) lines, indicating the fact that unnecessary responses are introduced by the regional matching with larger patch sizes. On the other hand, the bottom three examples of Figure~\ref{fig:statistic_compare} (B)  demonstrates the output $\mathcal{R}_{\psi,\theta}$ of the rectification module $\Theta$, and the noisy responses caused by irrelevant surrounding regions are suppressed by $\mathcal{R}_{\psi,\theta}$. 
Put differently, the input of $\Theta$ reflects how much each element of the support sample correlates with the ones in query sample, so it can be deemed as a kind of class-agnostic spatial distribution regarding the query-support activation.

\begin{figure}
\centering
    \begin{minipage}   {1.0\linewidth}
        \centering
        \includegraphics [width=1\linewidth] 
        {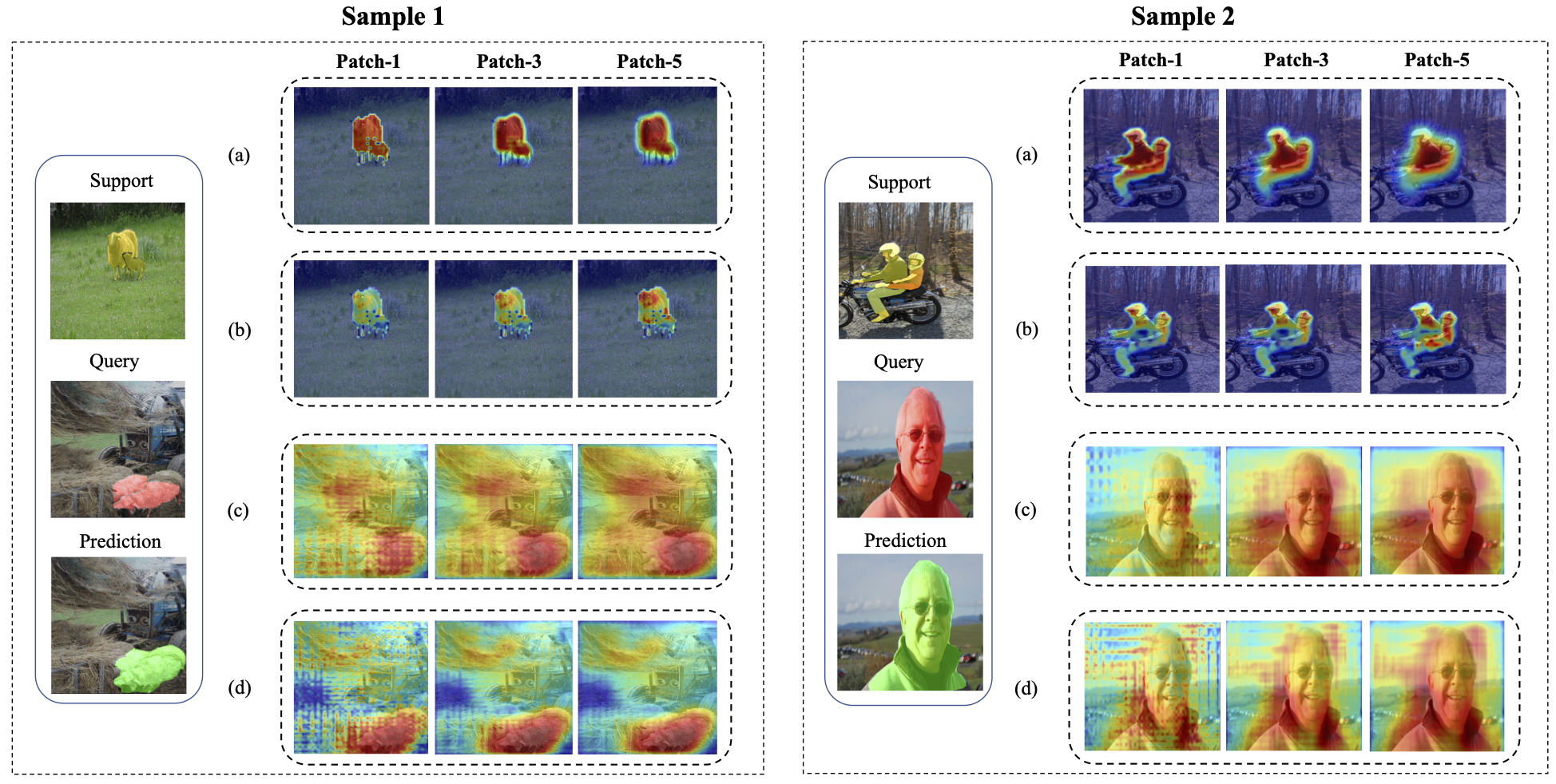}
    \end{minipage}
    \vspace{-0.2cm}    
    \caption{ 
    Qualitative visualizations of (a) $\mathcal{R}_{\psi}$, (b) $\mathcal{R}_{\psi,\theta}$, (c) the vanilla prior mask of PFENet and (d) the context-aware prior mask $\mathcal{Y}_{q, nf}$.}  
    \label{fig:weight}
\end{figure}

\pamiparagraph{More visual illustrations. }
The corresponding qualitative illustrations are presented in Figure~\ref{fig:weight} where (a), (b), (c) and (d) are the visualizations of $\mathcal{R}_{\psi}$, $\mathcal{R}_{\psi,\theta}$, the original prior mask of PFENet~\cite{pfenet} and the context-aware prior mask $\mathcal{Y}_{q, nf}$ adopted by PFENet++,  respectively. It can be observed in (b) that the importance of each support feature is highlighted by the rectifying factor $\mathcal{R}_{\psi,\theta}$, so that the context-aware prior masks shown in (d) carry more informative cues than that in (c).

\section{Boosting FSS with the Noise-filtered Context-aware Prior Mask}
Despite the fact that PFENet has shown its superiority by significantly outperforming its concurrent methods, as discussed in Sec.~\ref{sec:capm}, the original prior mask generation method in PFENet suffers from being context-imperceptive and noise-sensitive as it only considers the one-to-one relations without properly eliminating the adverse effects caused by the noisy support features. Therefore, in this section, we first briefly revisit PFENet and then propose PFENet++ by taking the essences of CAPM and NSM.

\subsection{Revisit PFENet }
To achieve promising few-shot segmentation performance without sinking into the over-fitting issues, CANet~\cite{canet} only concatenates the middle-level query and support features for yielding final predictions since they constitute object parts shared by unseen classes.
Different from CANet, Prior Guided Feature Enrichment Network (PFENet)~\cite{pfenet} leverages the middle- and high-level features separately to offer distinct types of guidance that are complementary to each other.

\begin{figure*}[!t]
\centering
    \begin{minipage}   {0.99\linewidth}
        \centering
        \includegraphics [width=1\linewidth] 
        {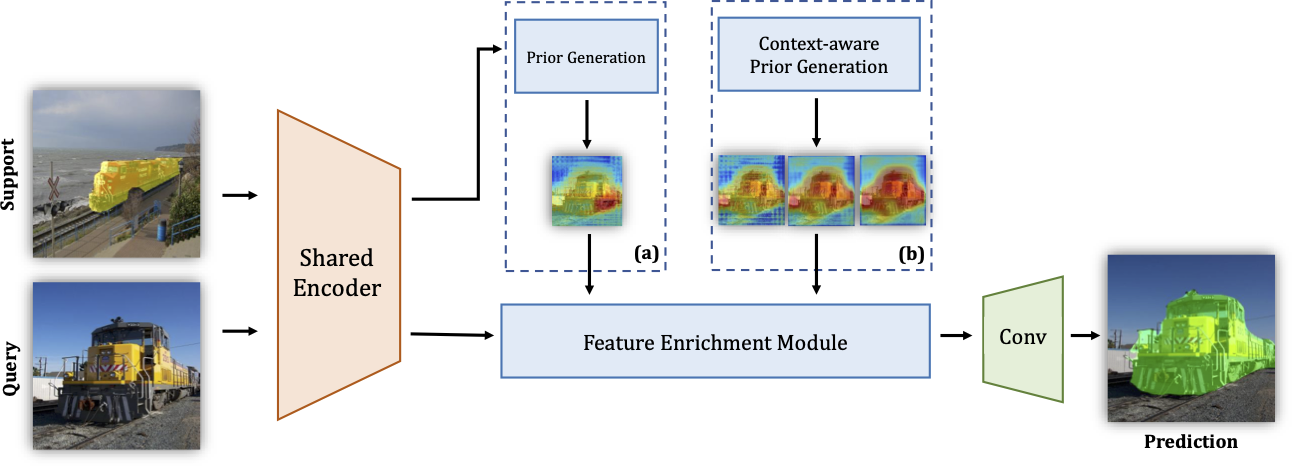}
    \end{minipage}
    \vspace{-0.2cm}
    \caption{The difference between PFENet~\cite{pfenet} and PFENet++ lies in their prior generation methods depicted in dashed boxes (a) and (b) respectively. The original prior generation method (a) only takes the maximum values from the one-to-one correspondence, while our new pipeline (b) yields preferable prior masks by decently incorporating additional contextual information with the proposed two sub-modules, \ie, CAPM and NSM. Besides, (a) only leverages the high-level features to yield prior masks, while (b) makes use of both high- and middle-level features. Adding middle-level features to (a) does not boost its performance.  }
    \label{fig:pfenet++}
\end{figure*}

The abstract structure of PFENet is shown in Fig.~\ref{fig:pfenet++} (a) where two modules are adopted, \ie, the Prior Generation Module (PGM) and Feature Enrichment Module (FEM). 
Specifically, PGM takes the high-level query and support features and it yields a single prior mask to indicate the region of interest on the query sample as introduced in Sec.~\ref{sec:prior_mask}. Then the prior mask is processed together with the middle-level query and support features by FEM where three steps, \ie, inter-source enrichment, inter-scale interaction and information concentration, are performed to yield robust predictions for the query images. More details about PFENet can be found in~\cite{pfenet}.

\subsection{PFENet++ }
As shown in Fig.~\ref{fig:pfenet++} (b), PFENet++ generally follows the design principles of PFENet, while the main upgrade lies in the prior generation process. PFENet++ operates with the proposed novel designs, \ie, Context-aware Prior Mask (CAPM) and Noise Suppression Module (NSM), to make the prior mask become more informative (\ie, context-aware) and resilient to noises. Similar to PFENet, prior masks generated by CAPM and NSM are then concatenated with the middle-level query and support features and processed by FEM to get the final prediction.

\pamiparagraph{Not only high-level features can offer prior guidance.}
It is worth noting that, in Fig.~\ref{fig:pfenet++} (b), the middle-level features in PFENet++ are made better use for offering extra prior guidance.
As verified in the empirical study of~\cite{pfenet}, the original prior mask generation method of PFENet yields inferior performance with the middle-level features, thus PFENet only exploits the correlation between the high-level features to yield the prior mask. However, we find that applying CAPM and NSM on the middle-level features on PFENet++ can bring additional performance gain to the model that has been incorporated with the prior masks obtained from the high-level features. We believe that it can be attributed to the noise filtering mechanism of NSM because the middle-level features are less semantic-sensitive and so are the correlations. If there is no selection performed on the middle-level support features, simply taking the maximum correspondence values among the correlation matrix might cause misleading responses on the prior masks. 

\pamiparagraph{Regional matching in CAPM.}
We empirically find that the set of patch sizes $M = \{1, 3, 5\}$ for regional matching yields decent performance by providing essential contextual information in CAPM, while the larger patch sizes, \eg, 7, cannot bring further improvement because the patch size 5 might have provided sufficient contextual aid. 
Since performing dense regional matching with three patch sizes leads to extra computation overhead compared to the original scheme $M$=$\{1\}$, the support feature map is thus down-sampled twice (\eg, $h_q \times w_q = 60 \times 60$, $h_s \times w_s = 30 \times 30$) for accomplishing regional matching in Eq.~\eqref{eqn:patch_similarity}. However, we observe that the down-sampled support feature map does not lead to performance deduction but retains satisfying efficiency. 

\pamiparagraph{Implementation of NSM.}
NSM involves two modules $\Psi$ and $\Theta$ for accomplishing local information compression and holistic rectification respectively. $\Psi$ is simply implemented by the averaging operation whose results are empirically found to be on par with that of the parameterized module. On the other hand,
$\Theta$ is composed of two fully-connected (FC) layers with an intermediate ReLU activation layer, and the weights of two FC layers are denoted as  $[h_s w_s \times d]$ and $[d \times h_s w_s]$. $\Theta$ first encodes the spatial information (\ie, correlation distribution) from $h_s w_s$ to $d$, followed by a decoding process from $d$ to $h_w w_s$ to produce $\mathcal{R}_{\psi,\theta} \in \mathbb{R}^{1 \times h_s w_s}$ that grasps the global context and then scores support features as in Eq.~\eqref{eqn:noise_filter_mask_generation}. 

\pamiparagraph{K-shot implementation.}
In the K-shot scenario, instead of forwarding the entire model for K times, we obtain K prior masks from K support samples and then we take the average of them as the final prior mask. Suppose if three scales are adopted by default, so we can get three averaged prior masks in total. Finally, same as PFENet, we concatenate these prior masks with the averaged middle-level features and pass them to FEM for yielding the final prediction. Thus only the feature extractor is additionally forwarded for K times, not the entire model. 

\section{Experiments}
\label{sec:results}

\begin{table*}
    \caption{Class mIoU and FB-IoU results on four folds of PASCAL-5$^i$. The results of `Mean' is the averaged class mIoU of four folds shown in the table. The detailed FB-IoU results of four folds are omitted in this table for simplicity. Models with $*$ adopt the original pre-trained ResNet~\cite{resnet}, and the ones without $*$ use the version optimized for semantic segmentation by following the classic PSPNet~\cite{pspnet}. } 
    \centering
    \tabcolsep=0.27cm
    {
        \begin{tabular}{ l |  c  c  c  c | c c |  c  c  c  c | c c  }
            \toprule
            \multirow{2}{*}{\textit{Methods}}  & \multicolumn{6}{c|}{1-Shot} & \multicolumn{6}{c}{5-Shot}   \\ 
            \specialrule{0em}{0pt}{1pt}
            \cline{2-13}
            \specialrule{0em}{1pt}{0pt} 
            & Fold-0 & Fold-1 & Fold-2 & Fold-3 & Mean & FB-IoU & Fold-0 & Fold-1 & Fold-2 & Fold-3 & Mean & FB-IoU \\
            
            \specialrule{0em}{0pt}{1pt}
            \hline
            \specialrule{0em}{1pt}{0pt}
            \multicolumn{13}{c}{VGG-16 Backbone} \\
            \specialrule{0em}{0pt}{1pt}
            \hline
            \specialrule{0em}{1pt}{0pt}
            
            OSLSM$_{2017}$ \cite{shaban} 
            & 33.6 & 55.3 & 40.9 & 33.5 & 40.8 & 61.3
            & 35.9 & 58.1 & 42.7 & 39.1 & 44.0 & 61.5
             \\  
            co-FCN$_{2018}$ \cite{co-FCN} 
            & 36.7 & 50.6 & 44.9 & 32.4 & 41.1 & 60.1
            & 37.5 & 50.0 & 44.1 & 33.9 & 41.4 & 60.2
             \\   
            SG-One$_{2018}$ \cite{SG-One} 
            & 40.2 & 58.4 & 48.4 & 38.4 & 46.3 & 63.9
            & 41.9 & 58.6 & 48.6 & 39.4 & 47.1 & 65.9
             \\     
            AMP$_{2019}$ \cite{adaptivemaskweightimprinting}
            & 41.9 & 50.2 & 46.7 & 34.7 & 43.4 & 61.9
            & 41.8 & 55.5 & 50.3 & 39.9 & 46.9 & 62.1
            \\
            PANet$_{2019}$ \cite{panet}
            & 42.3 & 58.0 & 51.1 & 41.2 & 48.1 & 66.5
            & 51.8 & 64.6 & 59.8 & 46.5 & 55.7 & 70.7
            \\
            FWBF$_{2019}$ \cite{weighingboosting} 
            & 47.0 & 59.6 & 52.6 & 48.3 & 51.9 & - 
            & 50.9 & 62.9 & 56.5 & 50.1 & 55.1 & -
            \\     
            RPMM$_{2020}$ \cite{protomix}  
            & 47.1 & 65.8 & 50.6 & 48.5 & 53.0 & -
            & 50.0 & 66.5 & 51.9 & 47.6 & 54.0 & -
             \\                 
    
            \specialrule{0em}{0pt}{1pt}
            \hline
            \specialrule{0em}{1pt}{0pt}
            PFENet$_{2020}$~\cite{pfenet} 
            & 56.9 & 68.2 & 54.4 & 52.4 &58.0 & 72.0
            & 59.0 & 69.1 & 54.8 & 52.9 &59.0 &72.3
            \\
            PFENet++  
            & 59.2 & 69.6 & 66.8 & 60.7 & \textbf{64.1} & \textbf{77.0}
            & 64.3 & 72.0 & 70.0 & 62.7 & \textbf{67.3} & \textbf{80.4}
            \\

            \specialrule{0em}{0pt}{1pt}
            \hline
            \specialrule{0em}{1pt}{0pt}
            \multicolumn{13}{c}{ResNet-50 Backbone} \\
            \specialrule{0em}{0pt}{1pt}
            \hline
            \specialrule{0em}{1pt}{0pt} 
            CANet$_{2019}^*$ \cite{canet}  
            & 52.5 & 65.9 & 51.3 & 51.9 & 55.4 & 66.2
            & 55.5 & 67.8 & 51.9 & 53.2 & 57.1 & 69.6
             \\      
  
            PPNet$_{2020}^*$\cite{ppnet}  
            & 48.6 & 60.6 & 55.7 &  46.5 & 52.8 & 69.2
            & 58.9 & 68.3 & 66.8 & 58.0 & 63.0 & 75.8
             \\ 
            RPMM$_{2020}^*$ \cite{protomix}  
            & 55.2 & 66.9 & 52.6 & 50.7 & 56.3 & -
            & 56.3 & 67.3 & 54.5 & 51.0 & 57.3 & -
             \\     

            PGNet$_{2019}$\cite{Zhang_2019_ICCV}  
            & 56.0 & 66.9 & 50.6 & 50.4 & 56.0 & 69.9
            & 54.9 & 67.4 & 51.8 & 53.0 & 56.8 & 70.5
             \\                    
            SCL$_{2021}$ \cite{scl} 
            & 63.0 & 70.0 & 56.5 & 57.7 & 61.8 & 71.9
            & 64.5 & 70.9 & 57.3 & 58.7 & 62.9 & 72.8
             \\           
            ASGNet$_{2021}$ \cite{asgnet} 
            & 58.8 & 67.9 & 56.8 & 53.7 & 59.3 & 69.2
            & 63.7 & 70.6 & 64.2 & 57.4 & 63.9 & 74.2
             \\         
            RePRI$_{2021}$ \cite{repri} 
            & 59.8 & 68.3 & 62.1 &  48.5 & 59.7 & -
            & 64.6 & 71.4 & 71.1 & 59.3 & 66.6 & -
             \\                    
            \specialrule{0em}{0pt}{1pt}
            \hline
            \specialrule{0em}{1pt}{0pt}            
            PFENet$_{2020}$~\cite{pfenet} 
            & 61.7 &69.5 & 55.4 & 56.3& 60.8 & 73.3
            & 63.1 & 70.7 & 55.8 &57.9 & 61.9& 73.9      
             \\   
            PFENet++ $^*$
            &60.6 & 70.3 & 65.6 & 60.3 &64.2 & 75.5& 65.2 & 73.6 & 74.1 & 65.3 & 69.6 & 80.8
            \\   
               PFENet++  
            & 63.3 & 71.0 & 65.9 & 59.6 & \textbf{64.9} & \textbf{76.8}
            & 66.1 & 75.0 & 74.1 & 64.3 & \textbf{69.9} & \textbf{81.1}
            \\

            \specialrule{0em}{0pt}{1pt}
            \hline
            \specialrule{0em}{1pt}{0pt}
            \multicolumn{13}{c}{ResNet-101 Backbone}  \\
            \specialrule{0em}{0pt}{1pt}
            \hline
            \specialrule{0em}{1pt}{0pt} 
            PPNet$_{2020}^*$ \cite{ppnet}  
            & 52.7 & 62.8 & 57.4 & 47.7 & 55.2 & 70.9
            & 60.3 & 70.0 & 69.4 & 60.7 & 65.1 & 77.5
             \\                  
            FWBF$_{2019}$ \cite{weighingboosting} 
            & 51.3 & 64.5 & 56.7 & 52.2 & 56.2 & -
            & 54.8 & 67.4 & 62.2 & 55.3 & 59.9 & -  
            \\            
            DAN$_{2020}$ \cite{democratic} 
            & 54.7 & 68.6 & 57.8 & 51.6 & 58.2 & 71.9
            & 57.9 & 69.0 & 60.1 & 54.9 & 60.5 & 72.3
            \\    
            ASGNet$_{2021}$ \cite{asgnet} 
            & 59.8 & 67.4 & 55.6 & 54.4 & 59.3 & 71.7 
            & 64.6 & 71.3 & 64.2 & 57.3 & 64.4 & 75.2
             \\                  
            RePRI$_{2021}$ \cite{repri} 
            & 59.6 & 68.6 & 62.2 & 47.2 & 59.4 & - 
            & 66.2 & 71.4 & 67.0 & 57.7 & 65.6 & -
             \\                   
            \specialrule{0em}{0pt}{1pt}
            \hline
            \specialrule{0em}{1pt}{0pt}            
            PFENet$_{2020}$~\cite{pfenet} 
            & 60.5 & 69.4 &54.4 &55.9 & 60.1 &72.9
            &62.8 &70.4& 54.9 &57.6 & 61.4& 73.5              
       \\   
            PFENet++ $^*$
            & 61.6 & 70.7 & 66.5 & 59.0 &64.5 & \textbf{76.3}
            & 64.3 & 74.9 & 73.9 & 66.3 & 69.9& 82.1
            \\   
               PFENet++
            & 63.1 & 72.4 & 63.4 & 62.2 & \textbf{65.3} & 75.5
            & 67.2 & 76.1 & 75.5 & 67.2 & \textbf{71.5} & \textbf{82.7}
             \\    

            \bottomrule            
        \end{tabular}
    }
    \label{tab:compare_sota_class}
\end{table*}

\begin{table*}
    \caption{Class mIoU and FB-IoU results on four folds of COCO-20$^i$. The results of `Mean' are the averaged class mIoU of four folds shown in the table. The detailed FB-IoU results of four folds are omitted in this table for simplicity. Models with $*$ adopt the original pre-trained ResNet~\cite{resnet}, and the ones without $*$ use the version optimized for semantic segmentation by following the classic PSPNet~\cite{pspnet}. } 
    \centering
    \tabcolsep=0.27cm
    {
        \begin{tabular}{ l |  c  c  c  c | c c |  c  c  c  c | c c  }
            \toprule
            \multirow{2}{*}{\textit{Methods}}  & \multicolumn{6}{c|}{1-Shot} & \multicolumn{6}{c}{5-Shot}   \\ 
            \specialrule{0em}{0pt}{1pt}
            \cline{2-13}
            \specialrule{0em}{1pt}{0pt} 
            & Fold-0 & Fold-1 & Fold-2 & Fold-3 & Mean & FB-IoU & Fold-0 & Fold-1 & Fold-2 & Fold-3 & Mean & FB-IoU \\
            
            \specialrule{0em}{0pt}{1pt}
            \hline
            \specialrule{0em}{1pt}{0pt}
            \multicolumn{13}{c}{VGG-16 Backbone} \\
            \specialrule{0em}{0pt}{1pt}
            \hline
            \specialrule{0em}{1pt}{0pt}
            
            PFENet$_{2020}$~\cite{pfenet} 
            & 33.4 & 36.0 & 34.1 & 32.8 & 34.1 & 60.0
            & 35.9 & 40.7 & 38.1 & 36.1 & 37.7 & 61.6
            \\
            PFENet++  
            & 38.6 & 43.1 & 40.0 & 39.5 & \textbf{40.3} &  \textbf{65.5}
            & 38.9 & 46.0 & 44.2 & 44.1 & \textbf{43.3} & \textbf{66.7}
            \\

            \specialrule{0em}{0pt}{1pt}
            \hline
            \specialrule{0em}{1pt}{0pt}
            \multicolumn{13}{c}{ResNet-50 Backbone} \\
            \specialrule{0em}{0pt}{1pt}
            \hline
            \specialrule{0em}{1pt}{0pt} 
  
            PPNet$_{2020}^*$ \cite{ppnet}  
            & 28.1 & 30.8 & 29.5 & 27.7 & 29.0 & -
            & 39.0 & 40.8 & 37.1 & 37.3 & 38.5 & -
             \\ 
               
           RPMM$_{2020}^*$ \cite{protomix}  
            & 29.5 & 36.8 & 28.9 & 27.0 & 30.6 & -
            & 33.8 & 42.0 & 33.0 & 33.3 & 35.5 & -
             \\                
               
            ASGNet$_{2021}$ \cite{asgnet} 
            & - & - & - & - & 34.6 & 60.4 
            & - & - & - & - & 42.5 & 67.0
             \\         
            RePRI$_{2021}$ \cite{repri} 
            & 31.2 & 38.1 & 33.3 & 33.0 & 34.0 & -
            & 38.5 & 46.2 & 40.0 & 43.6 & 42.1 & -
             \\   
          PFENet++ $^*$
            & 40.9 & 44.8 & 39.7 &38.8 &41.0 & 65.4&45.7& 52.4 & 49.1 & 47.2 & 48.6&69.4
            \\                   
            PFENet++  
            & 40.9 & 46.0 & 42.3 & 40.1 & \textbf{42.3} & \textbf{65.7}
            & 47.5 & 53.3 & 47.3 & 46.4 & \textbf{48.6} & \textbf{70.3}
            \\               
            		
            \specialrule{0em}{0pt}{1pt}
            \hline
            \specialrule{0em}{1pt}{0pt}
            \multicolumn{13}{c}{ResNet-101 Backbone}  \\
            \specialrule{0em}{0pt}{1pt}
            \hline
            \specialrule{0em}{1pt}{0pt} 
            FWBF$_{2019}$ \cite{weighingboosting} 
            & 17.0 & 18.0 & 21.0 & 28.9 & 21.2 & -
            & 19.1 & 21.5 & 23.9 & 30.1 & 23.7 & -
            \\            
            DAN$_{2020}$ \cite{democratic} 
            & - & - & - & - & 24.4 & 62.3
            & - & - & - & - & 29.6 & 63.9
            \\    
            SCL$_{2021}$ \cite{scl} 
            & 36.4 & 38.6 & 37.5 & 35.4 & 37.0 & -
            & 38.9 & 40.5 & 41.5 & 38.7 & 39.9 & -
             \\                        
            \specialrule{0em}{0pt}{1pt}
            \hline
            \specialrule{0em}{1pt}{0pt}            
            PFENet$_{2020}$~\cite{pfenet} 
            & 34.3 & 33.0 & 32.3 & 30.1 & 32.4 & 58.6
            & 38.5 & 38.6 & 38.2 & 34.3 & 37.4 & 61.9
                  \\    
             PFENet++  
            & 42.0 & 44.1 & 41.0 & 39.4 & \textbf{41.6} & \textbf{65.4}
            & 47.3 & 55.1 & 50.1 & 50.1 & \textbf{50.7} & \textbf{70.9}            
             \\   

            \bottomrule            
        \end{tabular}
    }
    \label{tab:compare_sota_class_coco}
\end{table*}

\begin{table}
    \caption{Foreground IoU results on FSS-1000~\cite{fss1000}.}
    \centering
    \tabcolsep=0.3cm
    {
        \begin{tabular}{ l |  c | c  c  }
            \toprule
            
            Methods  & 
            Backbone & 
            1-Shot & 
            5-Shot \\
            \specialrule{0em}{0pt}{1pt}
            \hline
            \specialrule{0em}{1pt}{0pt}
            OSLSM$_{2017}$~\cite{shaban} & \multirow{4}{*}{VGG-16} & 70.3 & 73.0 \\
            GN$_{2018}$~\cite{guide} & & 71.9 & 74.3 \\  
            FSS1000$_{2019}$~\cite{fss1000} & & 73.5 & 80.1 \\ 
            PFENet++ & & \textbf{86.5} & \textbf{87.5} \\ 
            \specialrule{0em}{0pt}{1pt}
            \hline
            \specialrule{0em}{1pt}{0pt}         
            PFENet++ &{ResNet-50}  & \textbf{88.6} & \textbf{89.1} \\
            \specialrule{0em}{0pt}{1pt}
            \hline
            \specialrule{0em}{1pt}{0pt}            
            DAN$_{2020}$~\cite{democratic} & \multirow{2}{*}{ResNet-101} & 85.2 & 88.1 \\  
             PFENet++ & & \textbf{88.6} & \textbf{89.2} \\           
            \bottomrule            
        \end{tabular}
    }
    \label{tab:compare_sota_fss1000}
\end{table}

\subsection{Implementation Details}
\label{sec:implementation_details}

\pamiparagraph{Datasets.}
The recent literature~\cite{canet,ppnet,pfenet,protomix} adopts three benchmarks PASCAL-5$^i$~\cite{shaban}, COCO-20$^i$~\cite{panet,weighingboosting} and FSS-1000~\cite{fss1000} for model evaluation in FSS. 

PASCAL-5$^i$ is constructed by combining PASCAL VOC
2012~\cite{pascalvoc2012} and extended annotations of SDS~\cite{SDS}. The cross-validation is performed by evenly dividing 20 classes into 4 folds $i \in \{0, 1, 2, 3\}$ and 5 classes in each fold. When evaluating one fold, the classes contained in the rest three folds are used as base classes for training.
Following~\cite{shaban,canet,panet,pfenet}, 1,000 query-support pairs are randomly sampled for the evaluation on each fold. For stability, we report the results averaged over 5 runs (\ie, 5,000 pairs).

COCO-20$^i$ is more much challenging than PASCAL-5$^i$ since the former contains 80 categories in total. Following~\cite{ppnet,repri,pfenet}, COCO-20$^i$ separates 80 classes from COCO~\cite{coco} in 4 folds and each fold contains 20 classes. It is observed in~\cite{pfenet} that evaluating COCO-20$^i$ with 1,000 episodes is far from enough since the validation set is much larger than that of PASCAL-5$^i$. The insufficient evaluation steps cause great performance variance as shown in~\cite{pfenet}, thus our models are instead evaluated with 20,000 episodes on COCO-20$^i$ for a fair comparison.

FSS-1000 contains 1,000 classes among which 486 classes are not included in any existing benchmarks and each class has 10 images with pixel-wise annotations. Following~\cite{fss1000}, the training, validation and test sets have 520, 240, and 240 classes respectively. We report the results obtained on the test set.

\pamiparagraph{Evaluation metrics.}
Following~\cite{canet,pfenet}, the class mean intersection over union (mIoU) is used as the main metric for comparison since the class mIoU is more informative than the foreground-background IoU (FB-IoU)~\cite{canet}. The calculation of class mIoU is: $mIoU = \frac{1}{C}\sum_{i=1}^{C} IoU_{i}$,
where $C$ is the number of categories contained in each fold (e.g., 5 for PASCAL-5$^i$ and 20 for COCO ) and $IoU_{i}$ is the IoU result of class $i$. The results of FB-IoU are also included for a comprehensive comparison where only the foreground and background are considered as two categories ($C$=2). Differently, FSS-1000 only measures the foreground IoU. We take average results of all folds as the final class mIoU/FB-IoU results on PASCAL-5$^i$ and COCO-20$^i$, while the results of FSS-1000 are directly obtained from its test set.

\pamiparagraph{Experimental configurations.}
All experiments are conducted on PyTorch framework. VGG-16 \cite{vgg}, ResNet-50 \cite{resnet} and ResNet-101 \cite{resnet} are used as the backbones, and the configurations are the same as that of PFENet~\cite{pfenet}. SGD is adopted as our optimizer. The momemtum and weight decay are set to 0.9 and 0.0001 respectively. The `poly' policy  \cite{deeplab} decays the learning rate by multiplying $(1 - \frac{current_{iter}}{max_{iter}})^{power}$ where $power$ equals to 0.9 according to the training progress.

Same as~\cite{canet,pfenet}, all our models are trained on PASCAL-5$^i$ for 200 epochs with the initial learning rate 0.0025 and batch size 4. On COCO, models are trained for 60 epochs with learning rate 0.006 and batch size 16. FSS-1000 is trained for 100 epochs with initial learning rate 0.01 and batch size 16. The ImageNet pre-trained backbone is kept unchanged during training. Data augmentation is important for combating over-fitting problems. Training samples are first randomly scaled from 0.9 to 1.1 and then randomly rotated from -10 to 10 degrees followed by the random mirroring operation with probability 0.5. After that, image patches are randomly cropped ($473\times473$ for PASCAL-5$^i$ and COCO-20$^i$, and $225\times225$ for FSS-1000) as the final training samples.

Since different datasets have various input image sizes, the hidden dimension number $d$ of $\Theta$ is set to 256 for PASCAL-5$^i$ and COCO-20$^i$, and 64 for FSS-1000. Similarly, the output dimension varies in different backbone networks, while we simply project all of them to 256, including the high-level features for yielding the prior masks for speeding up the prior generation process.

Additional post-processing strategies (\eg, multi-scale testing and DenseCRF \cite{densecrf}) are not implemented. Our experiments are performed on a single NVIDIA RTX 2080Ti GPU and Intel Xeon Silver 4216 CPU @ 2.10GHz. The implementation of PFENet++ follows the official repository\footnote{https://github.com/dvlab-research/PFENet} of PFENet~\cite{pfenet}.

\begin{figure*}[!t]
\centering
\resizebox{0.9\linewidth}{!}{
    \begin{tabular}{@{\hspace{0.5mm}}c@{\hspace{1.0mm}}c@{\hspace{1.0mm}}c@{\hspace{1.0mm}}c@{\hspace{2.0mm}}c@{\hspace{1.0mm}}c@{\hspace{1.0mm}}c@{\hspace{1.0mm}}c@{\hspace{0.0mm}}}
		\rotatebox[origin=c]{90}{\scriptsize{(a) Support}}
		\includegraphics[align=c,width=0.15\linewidth,height=0.12\linewidth]{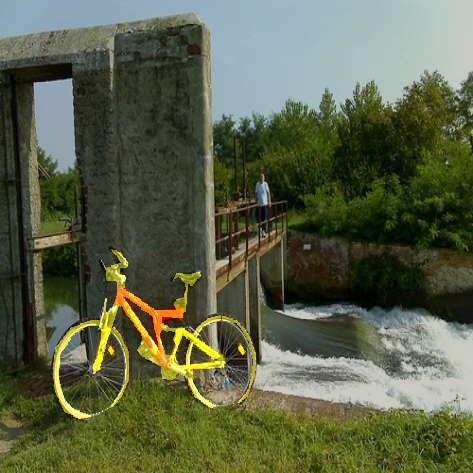}&
		\includegraphics[align=c,width=0.15\linewidth,height=0.12\linewidth]{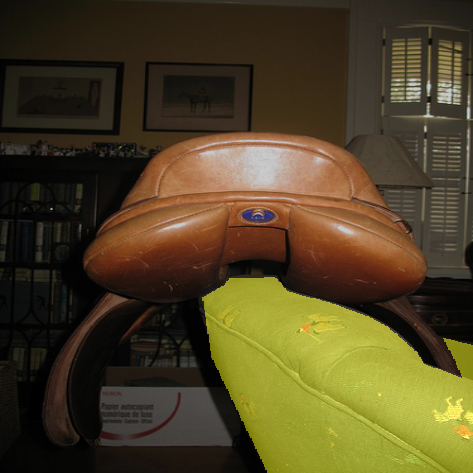}&
		\includegraphics[align=c,width=0.15\linewidth,height=0.12\linewidth]{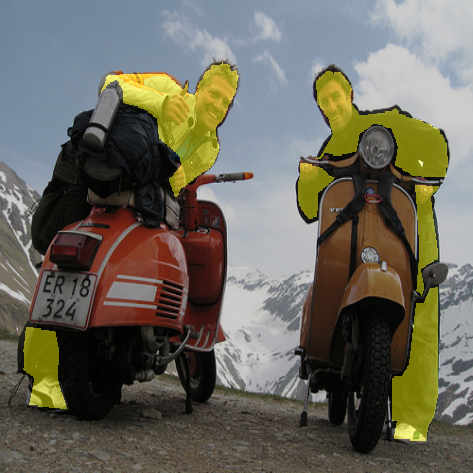}&
		\includegraphics[align=c,width=0.15\linewidth,height=0.12\linewidth]{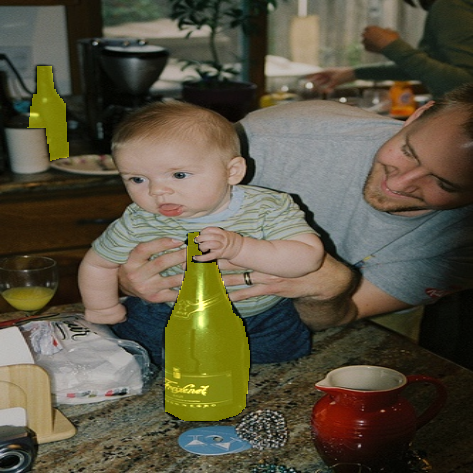}&
		\includegraphics[align=c,width=0.15\linewidth,height=0.12\linewidth]{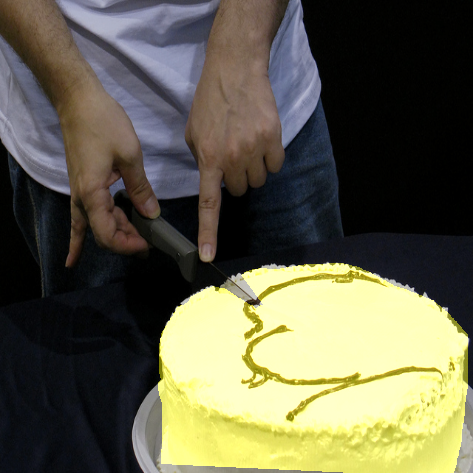}&
		\includegraphics[align=c,width=0.15\linewidth,height=0.12\linewidth]{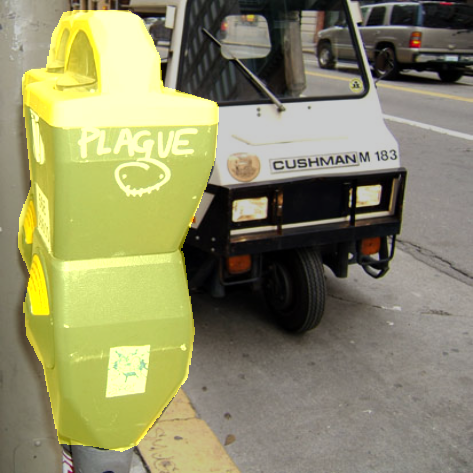}&
		\includegraphics[align=c,width=0.15\linewidth,height=0.12\linewidth]{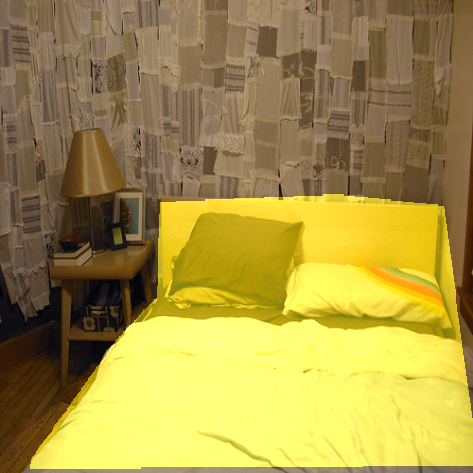}&
		\includegraphics[align=c,width=0.15\linewidth,height=0.12\linewidth]{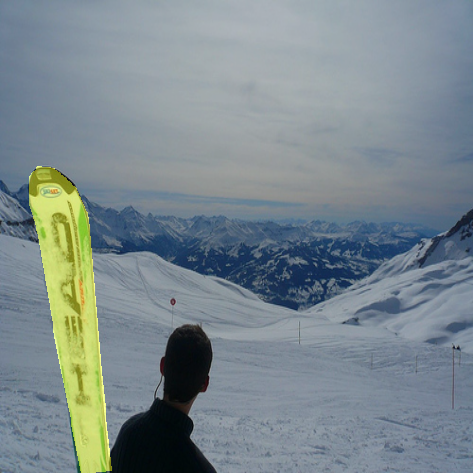}\\
		\addlinespace[3pt]
		\rotatebox[origin=c]{90}{\scriptsize{(b) Query}}
		\includegraphics[align=c,width=0.15\linewidth,height=0.12\linewidth]{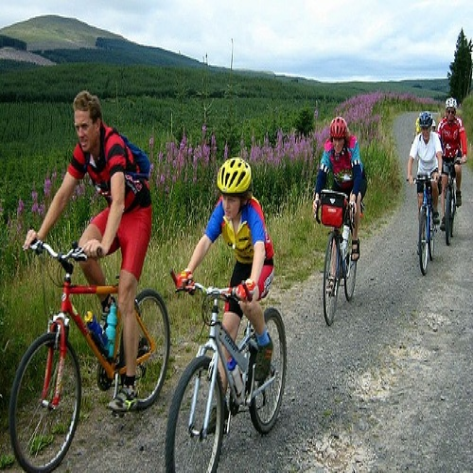}&
		\includegraphics[align=c,width=0.15\linewidth,height=0.12\linewidth]{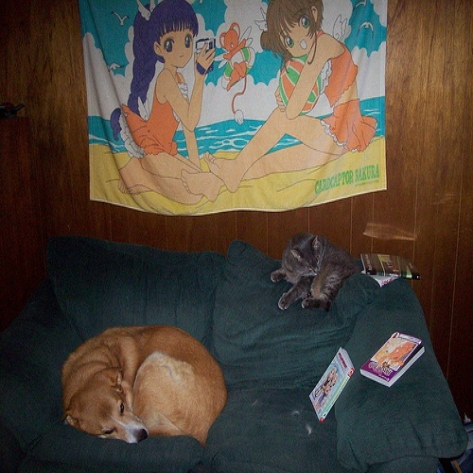}&
		\includegraphics[align=c,width=0.15\linewidth,height=0.12\linewidth]{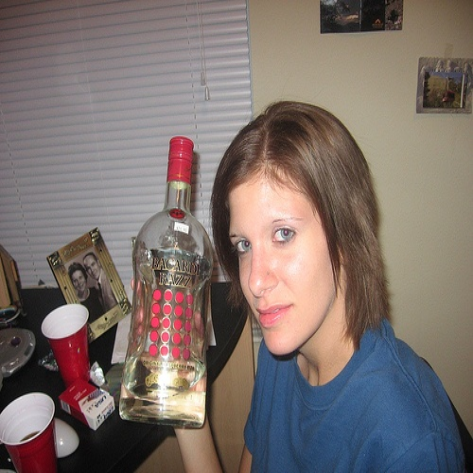}&
		\includegraphics[align=c,width=0.15\linewidth,height=0.12\linewidth]{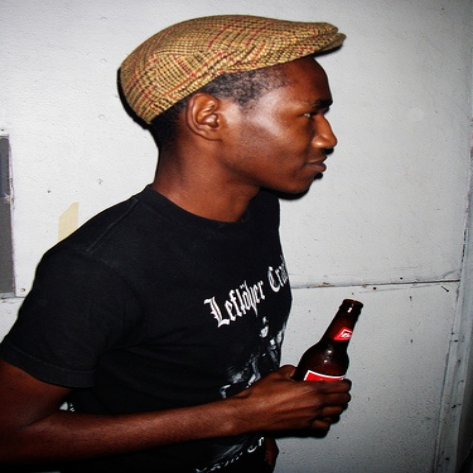}&
		\includegraphics[align=c,width=0.15\linewidth,height=0.12\linewidth]{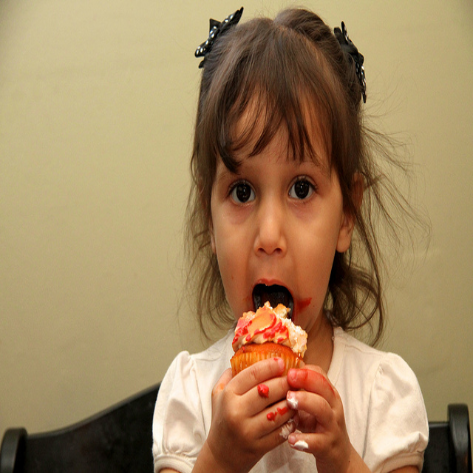}&
		\includegraphics[align=c,width=0.15\linewidth,height=0.12\linewidth]{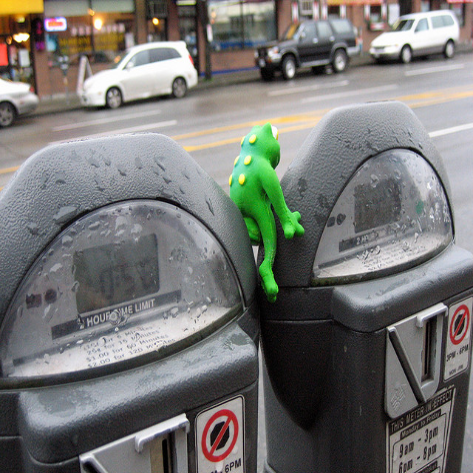}&
		\includegraphics[align=c,width=0.15\linewidth,height=0.12\linewidth]{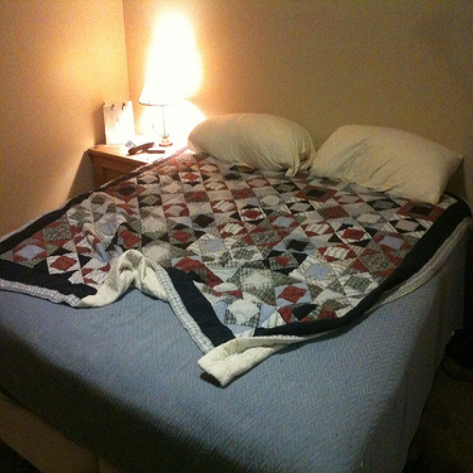}&
		\includegraphics[align=c,width=0.15\linewidth,height=0.12\linewidth]{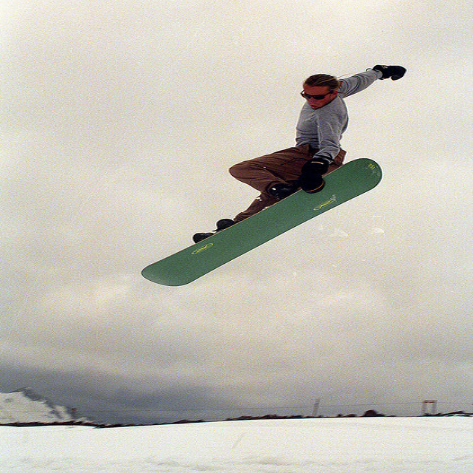}\\
		\addlinespace[3pt]
		\rotatebox[origin=c]{90}{\scriptsize{(c) GT}}
		\includegraphics[align=c,width=0.15\linewidth,height=0.12\linewidth]{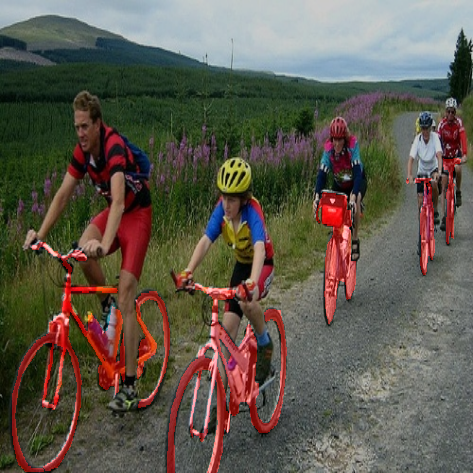}&
		\includegraphics[align=c,width=0.15\linewidth,height=0.12\linewidth]{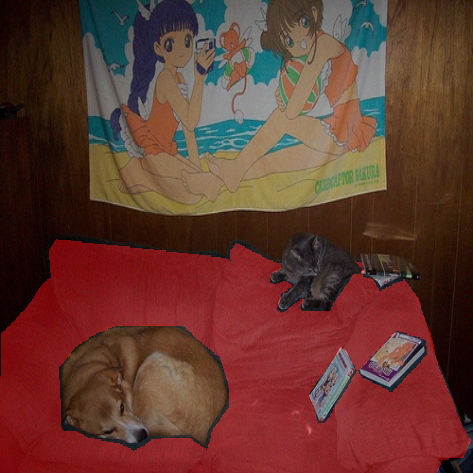}&
		\includegraphics[align=c,width=0.15\linewidth,height=0.12\linewidth]{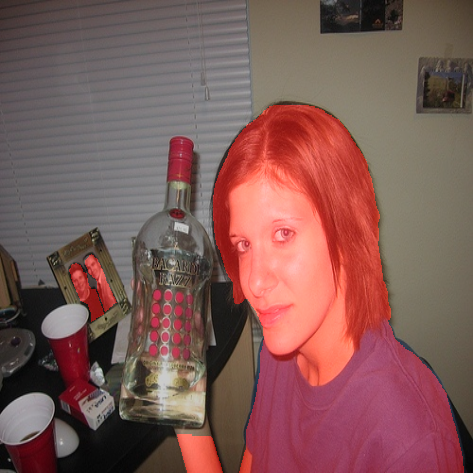}&
		\includegraphics[align=c,width=0.15\linewidth,height=0.12\linewidth]{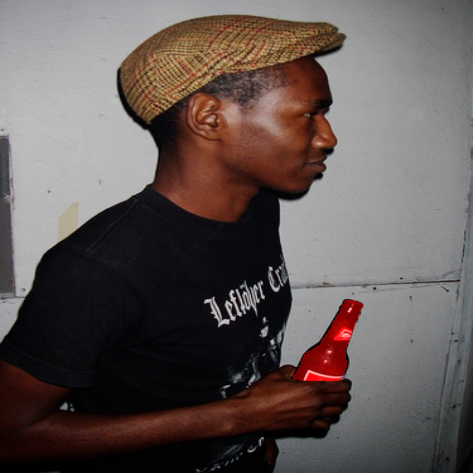}&
		\includegraphics[align=c,width=0.15\linewidth,height=0.12\linewidth]{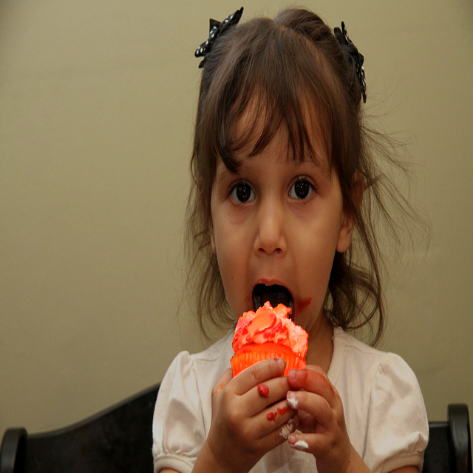}&
		\includegraphics[align=c,width=0.15\linewidth,height=0.12\linewidth]{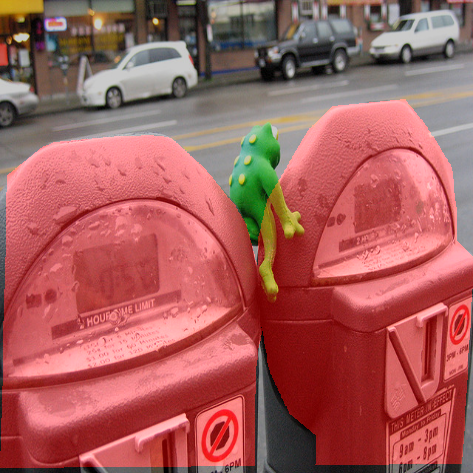}&
		\includegraphics[align=c,width=0.15\linewidth,height=0.12\linewidth]{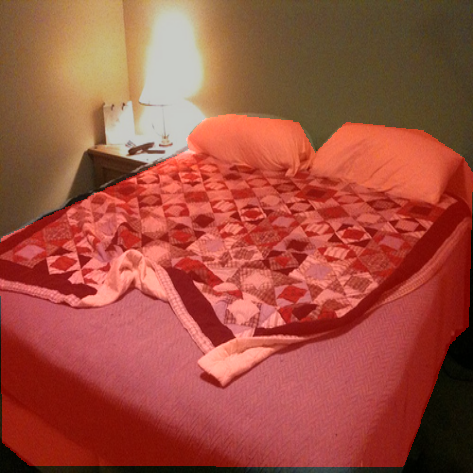}&
		\includegraphics[align=c,width=0.15\linewidth,height=0.12\linewidth]{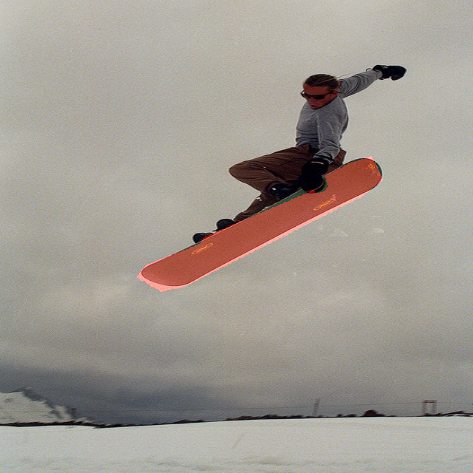}\\
		\addlinespace[3pt]
		\rotatebox[origin=c]{90}{\scriptsize{(d) PFENet}}
		\includegraphics[align=c,width=0.15\linewidth,height=0.12\linewidth]{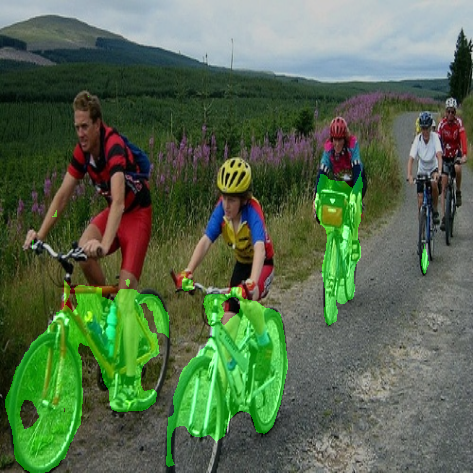}&
		\includegraphics[align=c,width=0.15\linewidth,height=0.12\linewidth]{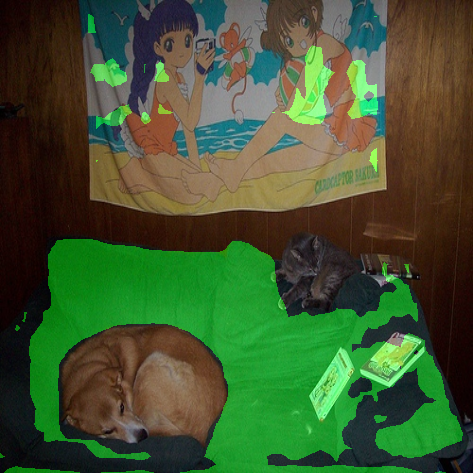}&
		\includegraphics[align=c,width=0.15\linewidth,height=0.12\linewidth]{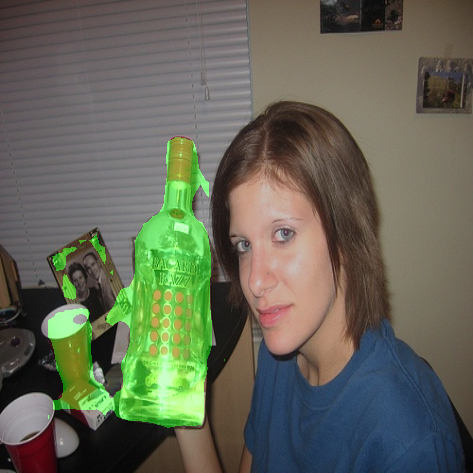}&
		\includegraphics[align=c,width=0.15\linewidth,height=0.12\linewidth]{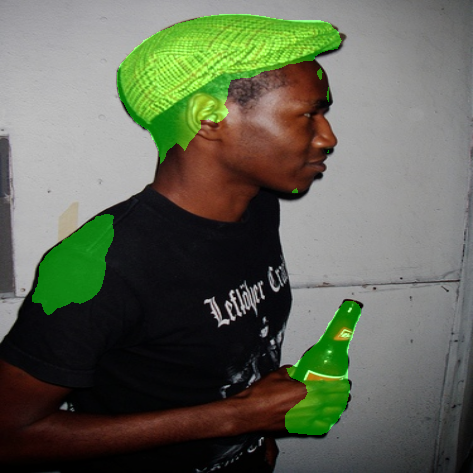}&
		\includegraphics[align=c,width=0.15\linewidth,height=0.12\linewidth]{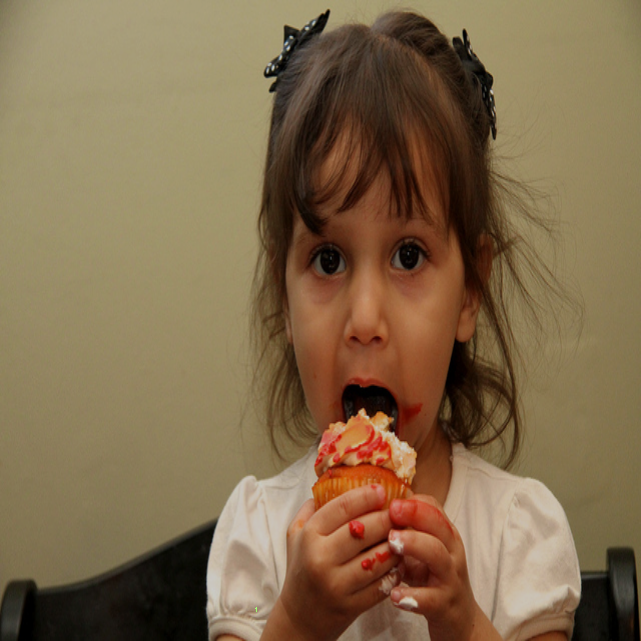}&
		\includegraphics[align=c,width=0.15\linewidth,height=0.12\linewidth]{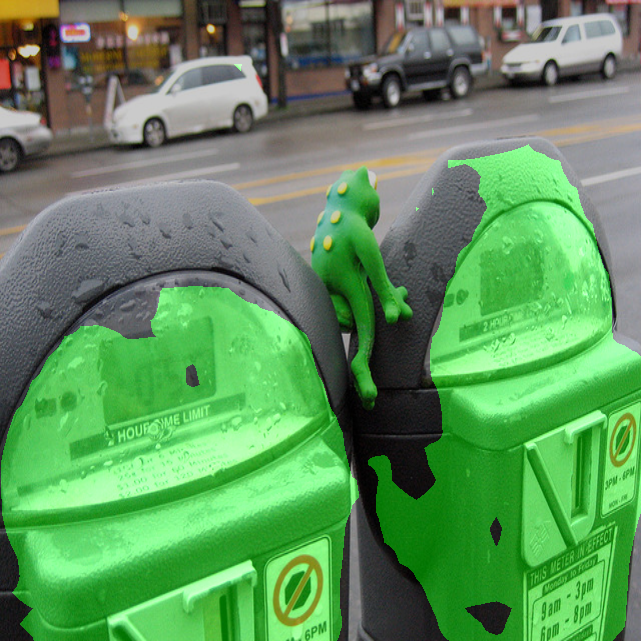}&
		\includegraphics[align=c,width=0.15\linewidth,height=0.12\linewidth]{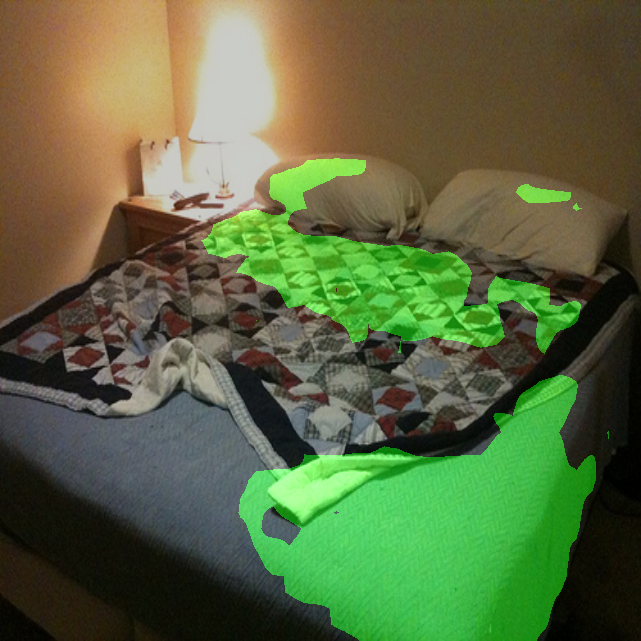}&
		\includegraphics[align=c,width=0.15\linewidth,height=0.12\linewidth]{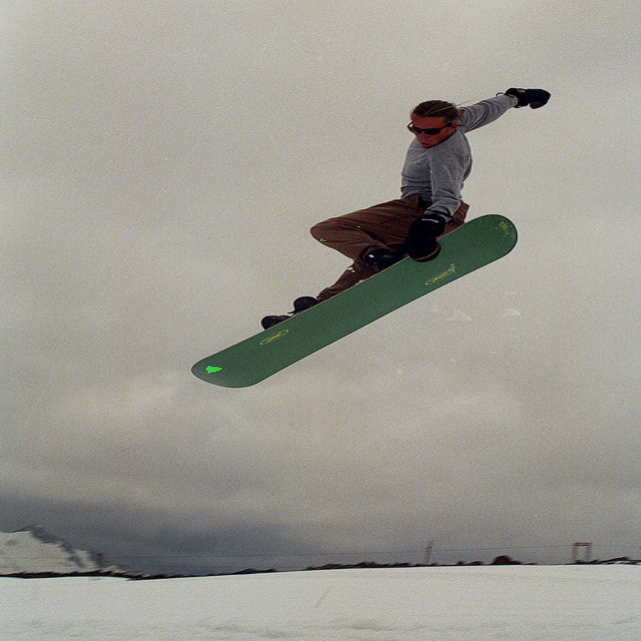}\\
		\addlinespace[3pt]
		\rotatebox[origin=c]{90}{\scriptsize{(e) Prior}}
		\includegraphics[align=c,width=0.15\linewidth,height=0.12\linewidth]{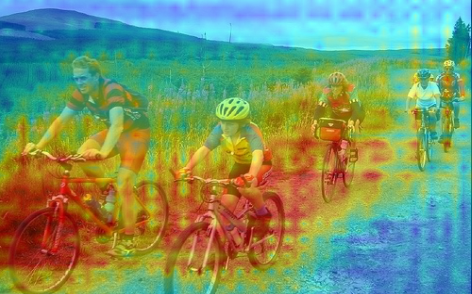}&
		\includegraphics[align=c,width=0.15\linewidth,height=0.12\linewidth]{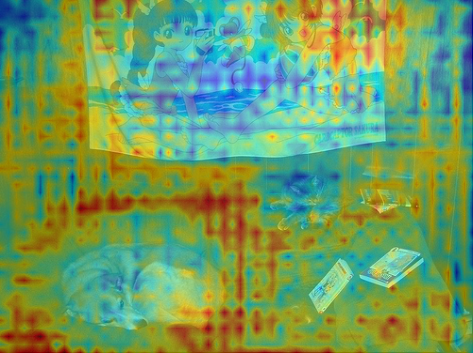}&
		\includegraphics[align=c,width=0.15\linewidth,height=0.12\linewidth]{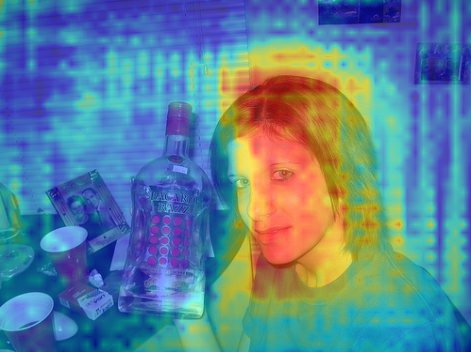}&
		\includegraphics[align=c,width=0.15\linewidth,height=0.12\linewidth]{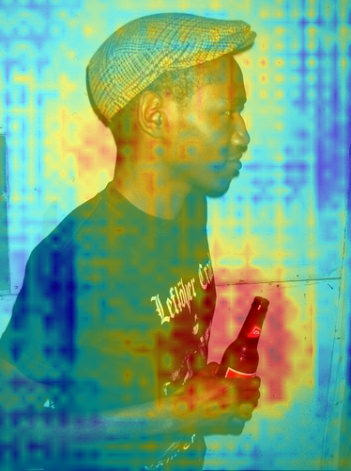}&
		\includegraphics[align=c,width=0.15\linewidth,height=0.12\linewidth]{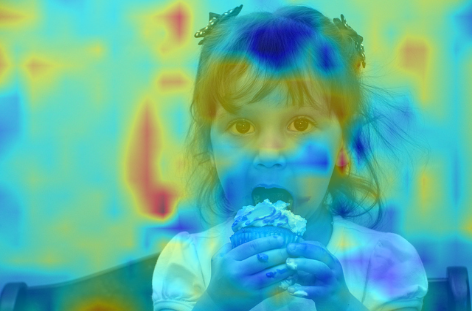}&
		\includegraphics[align=c,width=0.15\linewidth,height=0.12\linewidth]{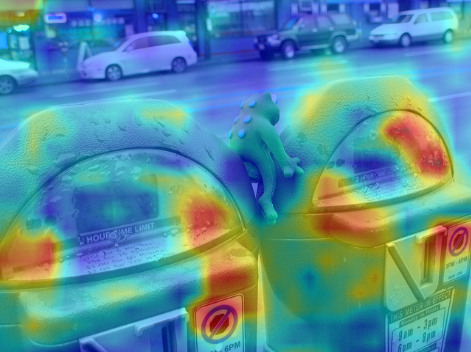}&
		\includegraphics[align=c,width=0.15\linewidth,height=0.12\linewidth]{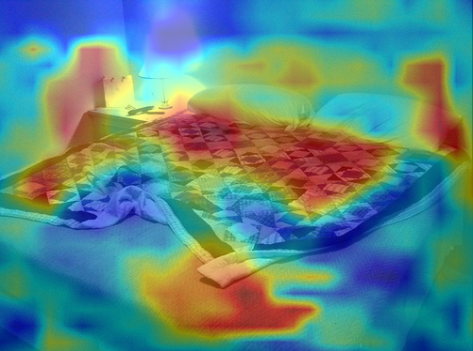}&
		\includegraphics[align=c,width=0.15\linewidth,height=0.12\linewidth]{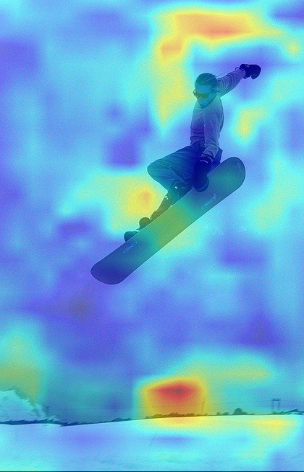}\\
		\addlinespace[3pt]
		\rotatebox[origin=c]{90}{\scriptsize{(f) PFENet++}}
		\includegraphics[align=c,width=0.15\linewidth,height=0.12\linewidth]{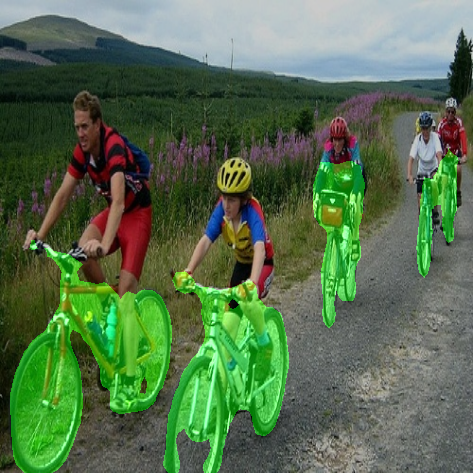}&
		\includegraphics[align=c,width=0.15\linewidth,height=0.12\linewidth]{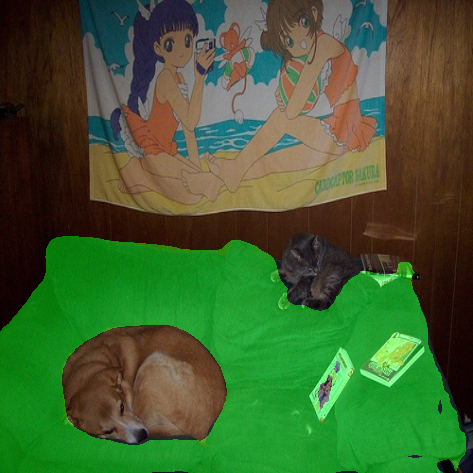}&
		\includegraphics[align=c,width=0.15\linewidth,height=0.12\linewidth]{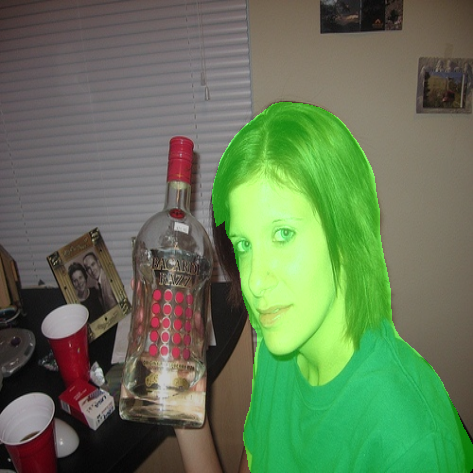}&
		\includegraphics[align=c,width=0.15\linewidth,height=0.12\linewidth]{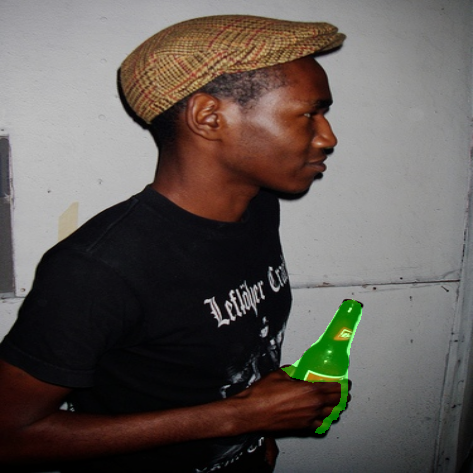}&
		\includegraphics[align=c,width=0.15\linewidth,height=0.12\linewidth]{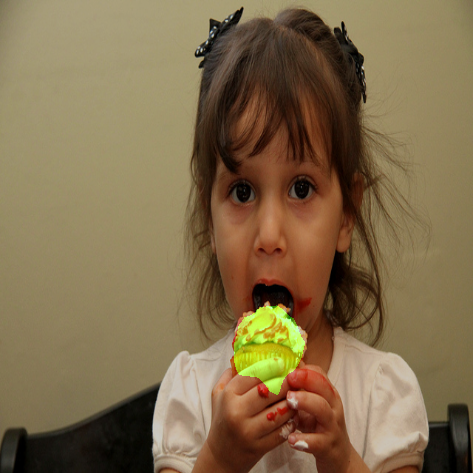}&
		\includegraphics[align=c,width=0.15\linewidth,height=0.12\linewidth]{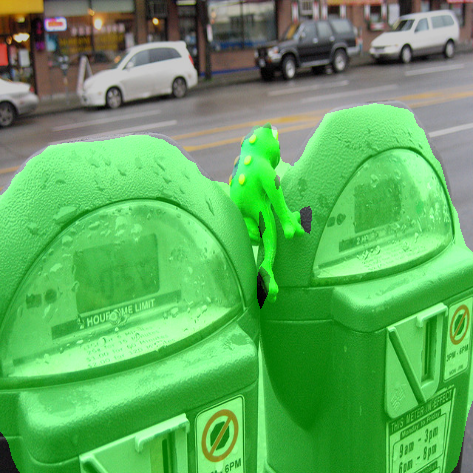}&
		\includegraphics[align=c,width=0.15\linewidth,height=0.12\linewidth]{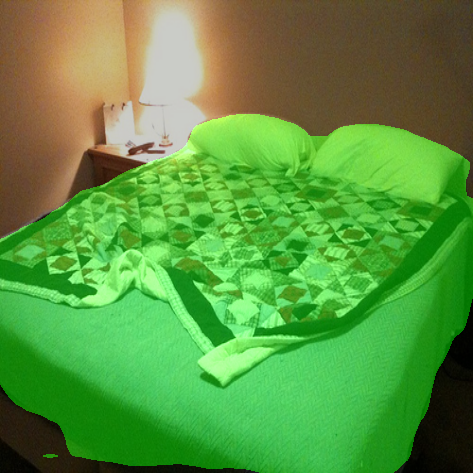}&
		\includegraphics[align=c,width=0.15\linewidth,height=0.12\linewidth]{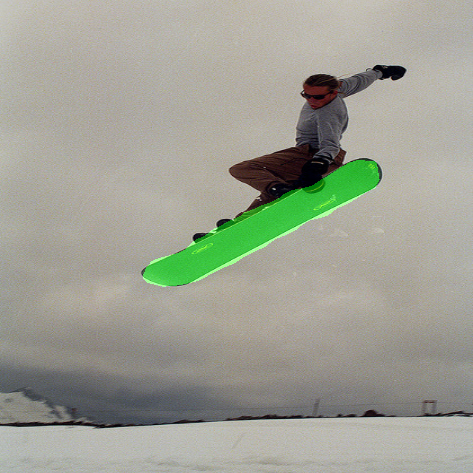}\\
		\addlinespace[3pt]
		\rotatebox[origin=c]{90}{\scriptsize{(g) Prior$_{1\times1}$}}
		\includegraphics[align=c,width=0.15\linewidth,height=0.12\linewidth]{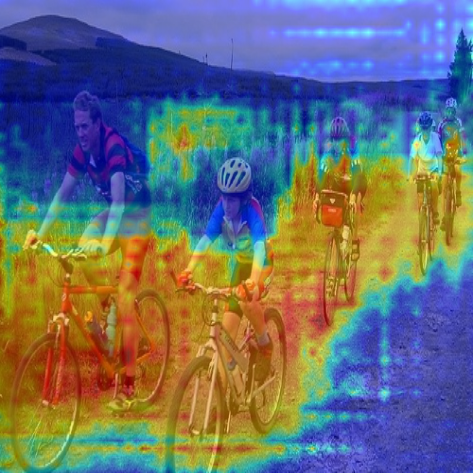}&
		\includegraphics[align=c,width=0.15\linewidth,height=0.12\linewidth]{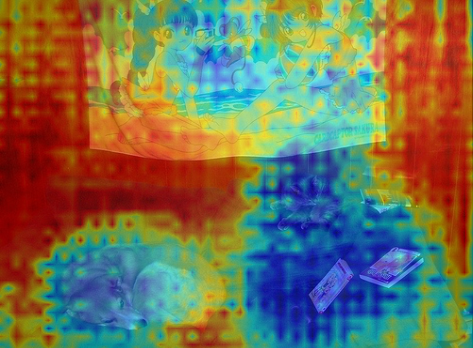}&
		\includegraphics[align=c,width=0.15\linewidth,height=0.12\linewidth]{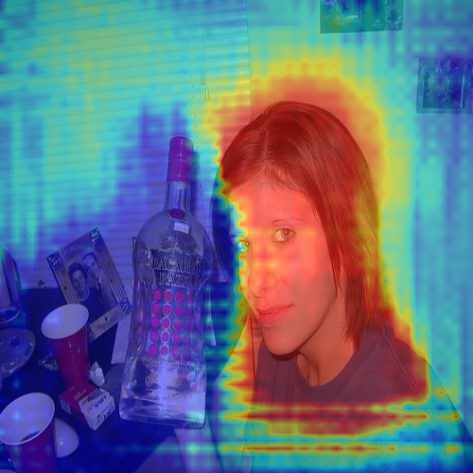}&
		\includegraphics[align=c,width=0.15\linewidth,height=0.12\linewidth]{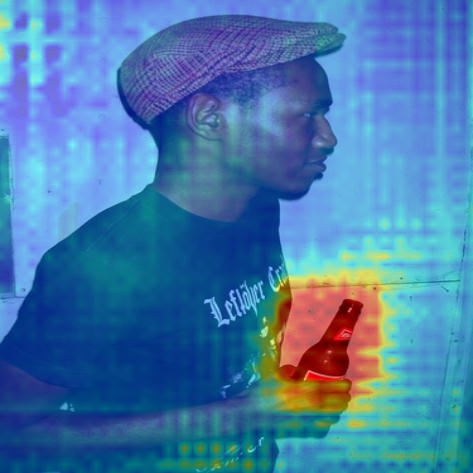}&
		\includegraphics[align=c,width=0.15\linewidth,height=0.12\linewidth]{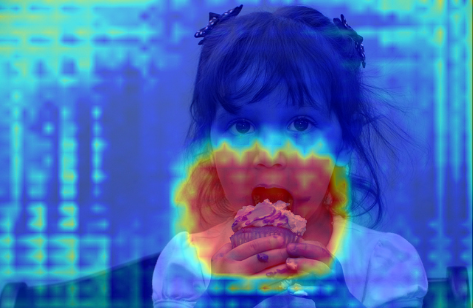}&
		\includegraphics[align=c,width=0.15\linewidth,height=0.12\linewidth]{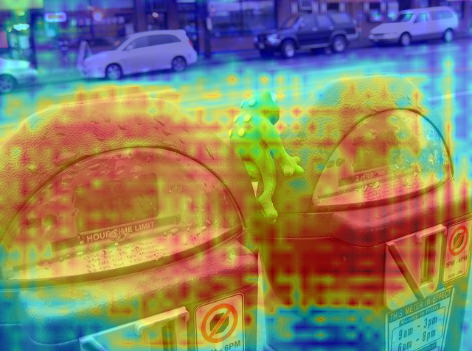}&
		\includegraphics[align=c,width=0.15\linewidth,height=0.12\linewidth]{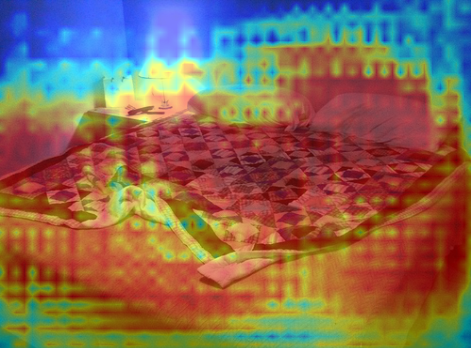}&
		\includegraphics[align=c,width=0.15\linewidth,height=0.12\linewidth]{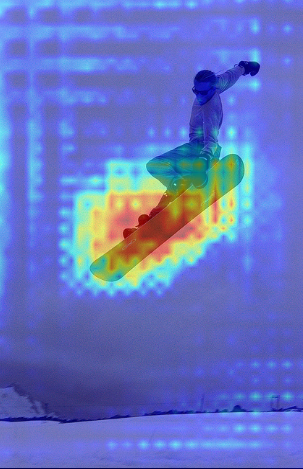}\\
		\addlinespace[3pt]
	    \rotatebox[origin=c]{90}{\scriptsize{(h) Prior$_{3\times3}$}}
		\includegraphics[align=c,width=0.15\linewidth,height=0.12\linewidth]{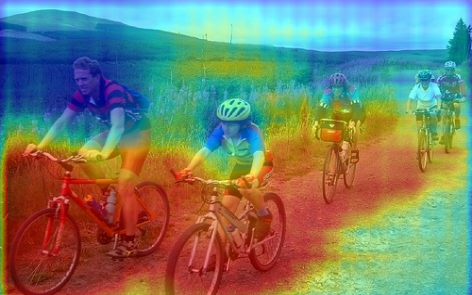}&
		\includegraphics[align=c,width=0.15\linewidth,height=0.12\linewidth]{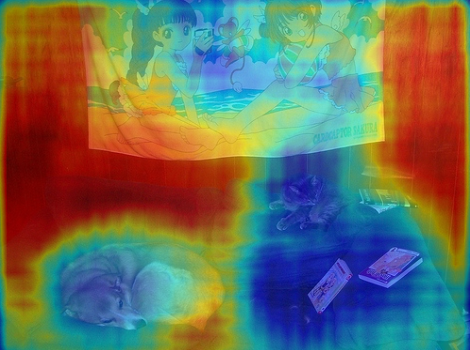}&
		\includegraphics[align=c,width=0.15\linewidth,height=0.12\linewidth]{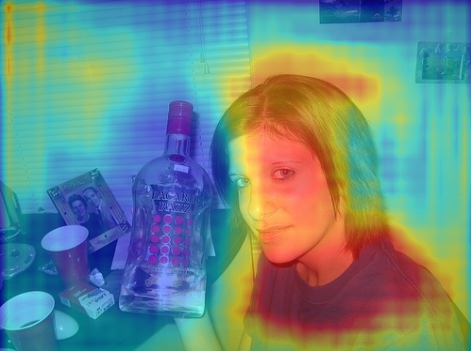}&
		\includegraphics[align=c,width=0.15\linewidth,height=0.12\linewidth]{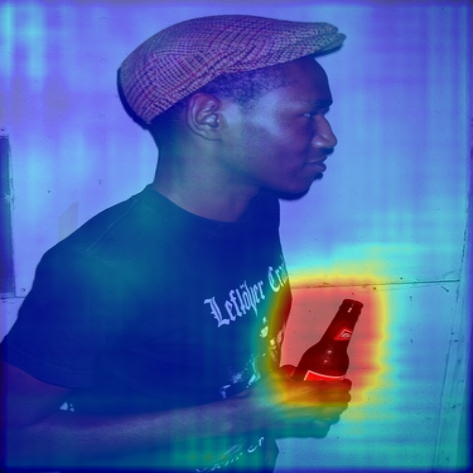}&
		\includegraphics[align=c,width=0.15\linewidth,height=0.12\linewidth]{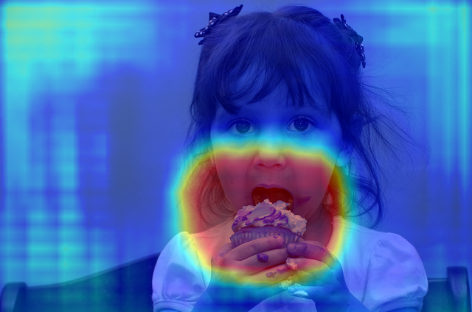}&
		\includegraphics[align=c,width=0.15\linewidth,height=0.12\linewidth]{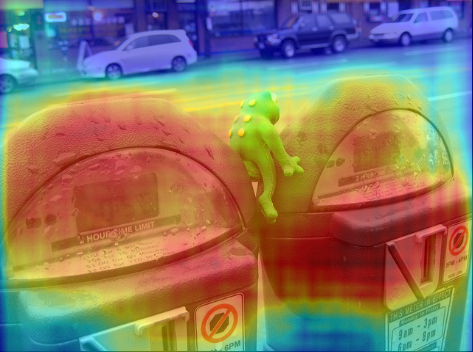}&
		\includegraphics[align=c,width=0.15\linewidth,height=0.12\linewidth]{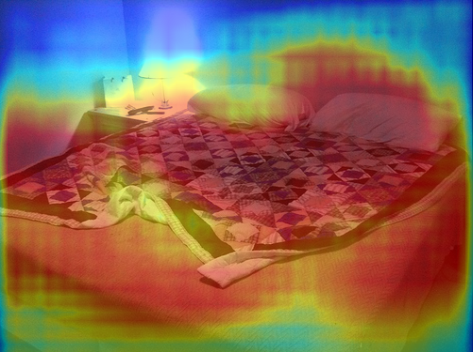}&
		\includegraphics[align=c,width=0.15\linewidth,height=0.12\linewidth]{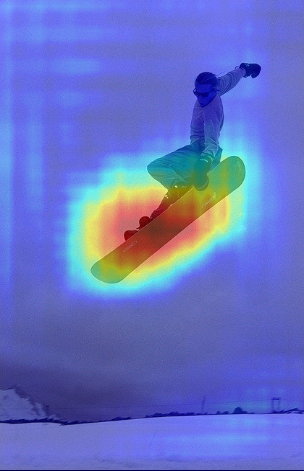}\\
		\addlinespace[3pt]
		\rotatebox[origin=c]{90}{\scriptsize{(i) Prior$_{5\times5}$}}
		\includegraphics[align=c,width=0.15\linewidth,height=0.12\linewidth]{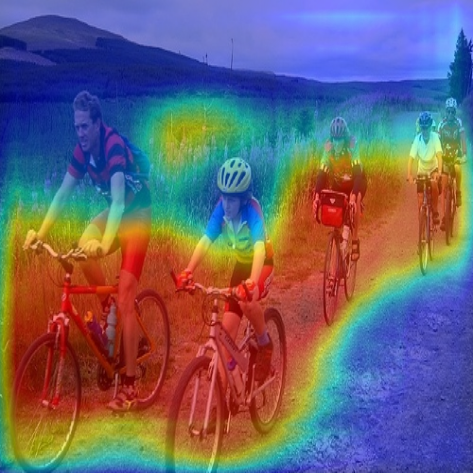}&
		\includegraphics[align=c,width=0.15\linewidth,height=0.12\linewidth]{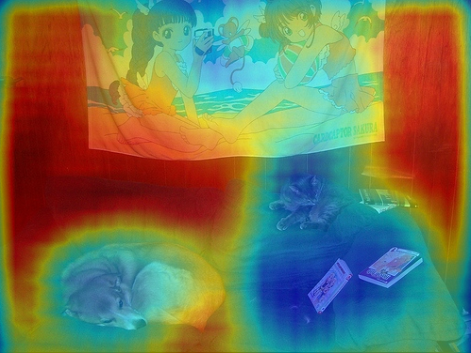}&
		\includegraphics[align=c,width=0.15\linewidth,height=0.12\linewidth]{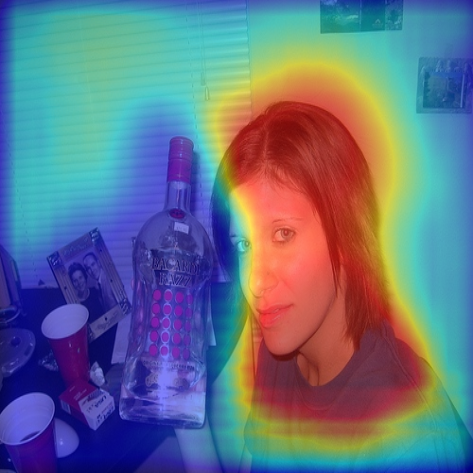}&
		\includegraphics[align=c,width=0.15\linewidth,height=0.12\linewidth]{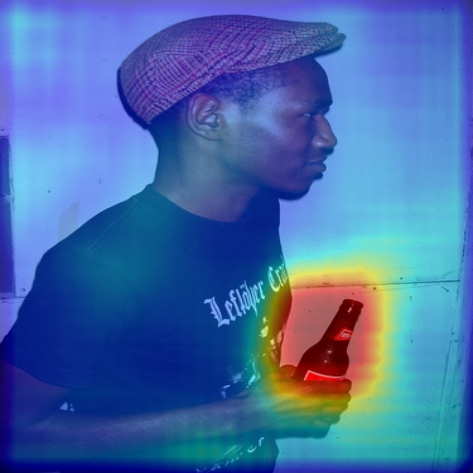}&
		\includegraphics[align=c,width=0.15\linewidth,height=0.12\linewidth]{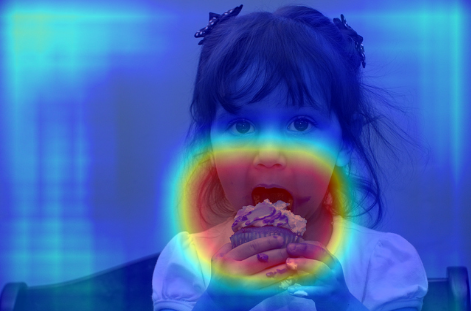}&
		\includegraphics[align=c,width=0.15\linewidth,height=0.12\linewidth]{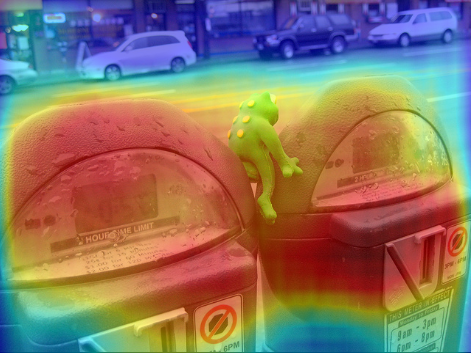}&
		\includegraphics[align=c,width=0.15\linewidth,height=0.12\linewidth]{coco_406_1_prior.png}&
		\includegraphics[align=c,width=0.15\linewidth,height=0.12\linewidth]{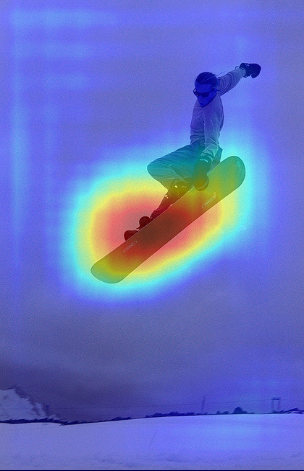}
    \end{tabular}}
\caption{Qualitative results of the proposed PFENet++ and the original PFENet. The right samples are from COCO and the left ones are from PASCAL-5$^i$. From top to bottom: (a) support images; (b) query images; (c) ground truth of query images; (d) predictions of PFENet; (e) the vanilla prior mask of PFENet; (f) the predictions of PFENet++; (g)(h)(i) are the proposed context-aware prior masks yielded by $1\times1$, $3\times3$, $5\times5$ patches, respectively. }
\label{fig:visual_compare}
\end{figure*}

\begin{figure*}
\centering

    \begin{minipage}   {0.49\linewidth}
        \centering
        \includegraphics [width=1\linewidth] 
        {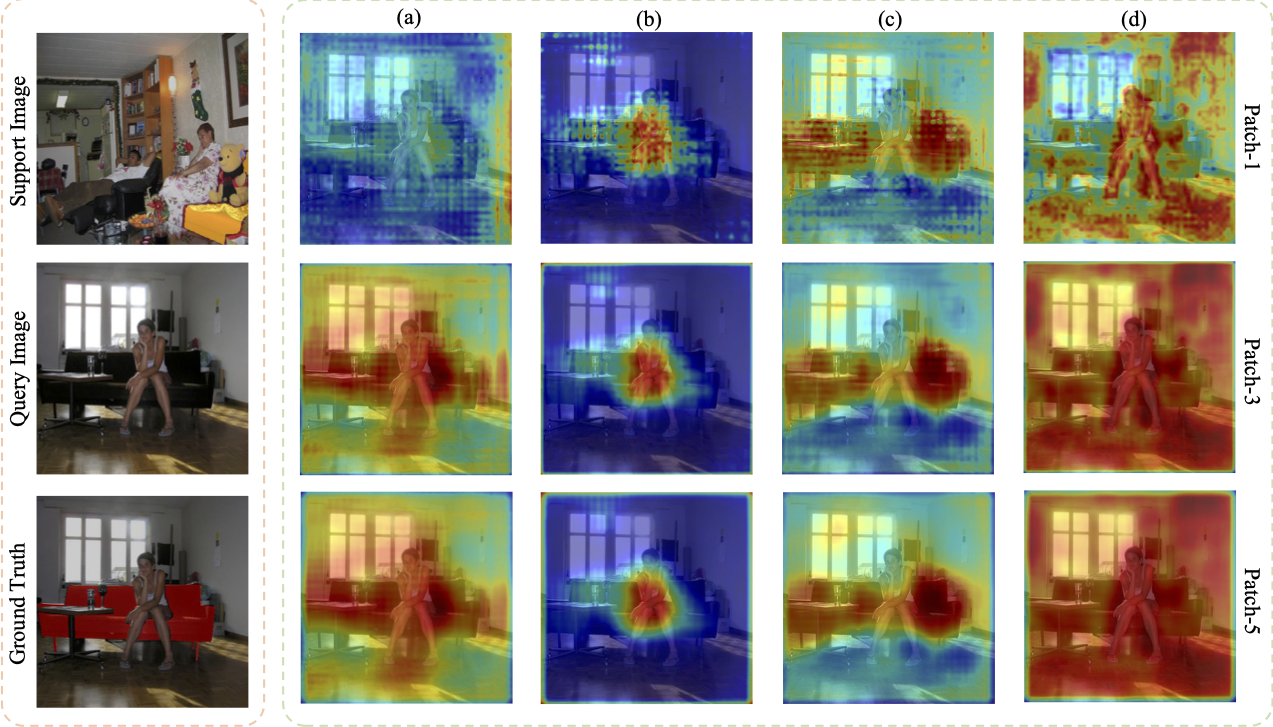}
    \end{minipage}
    \begin{minipage}   {0.49\linewidth}
        \centering
        \includegraphics [width=1\linewidth] 
        {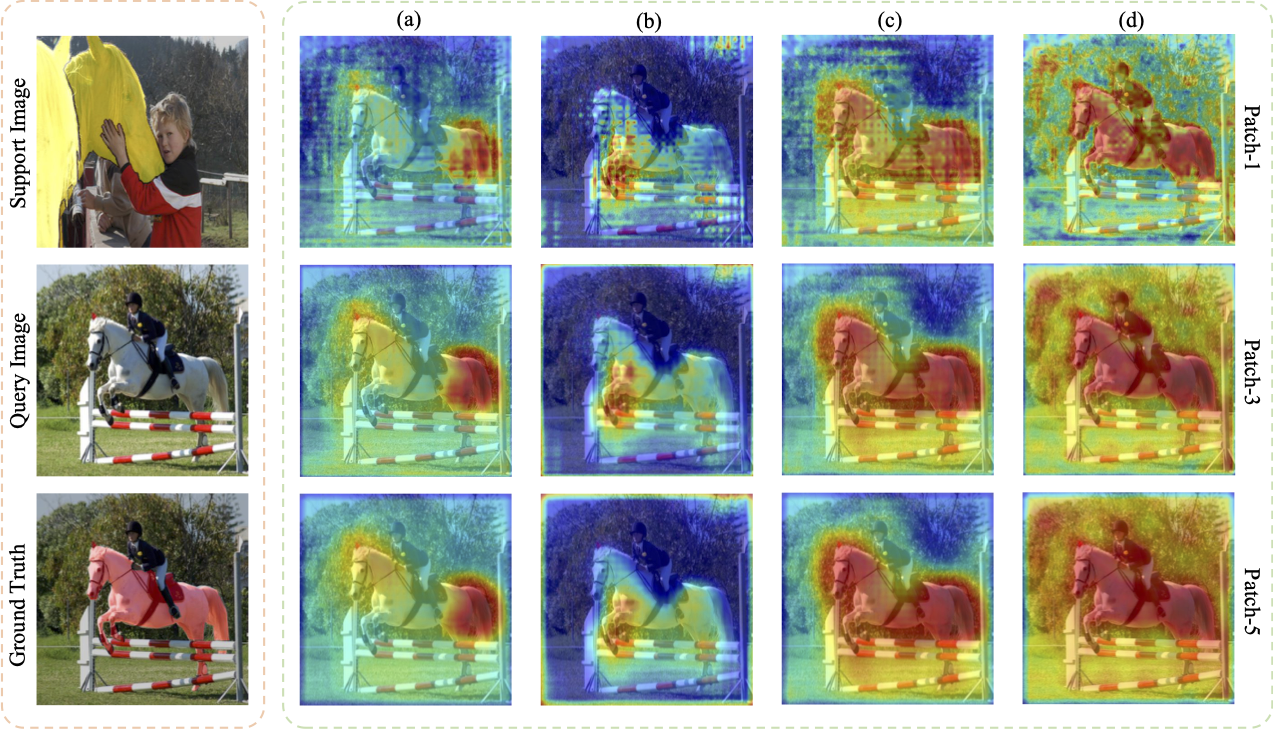}
    \end{minipage}   
    
    \begin{minipage}   {0.49\linewidth}
        \centering
        \includegraphics [width=1\linewidth] 
        {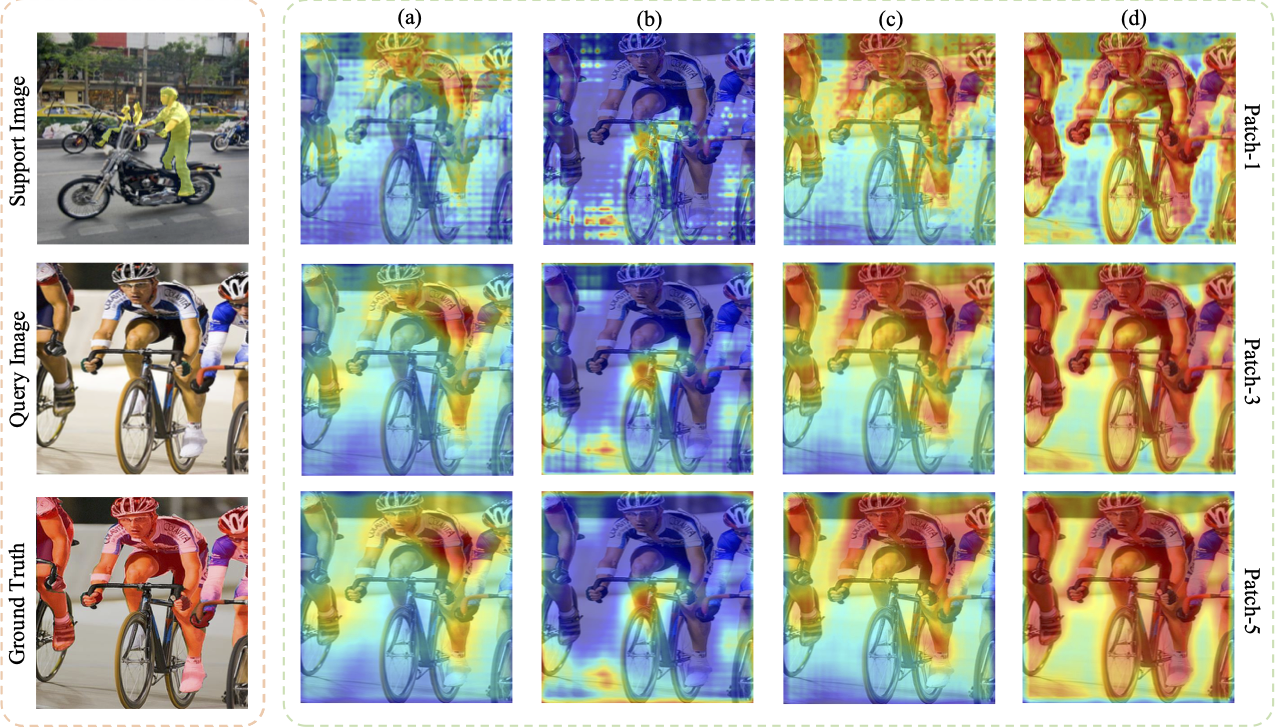}
    \end{minipage}
    \begin{minipage}   {0.49\linewidth}
        \centering
        \includegraphics [width=1\linewidth] 
        {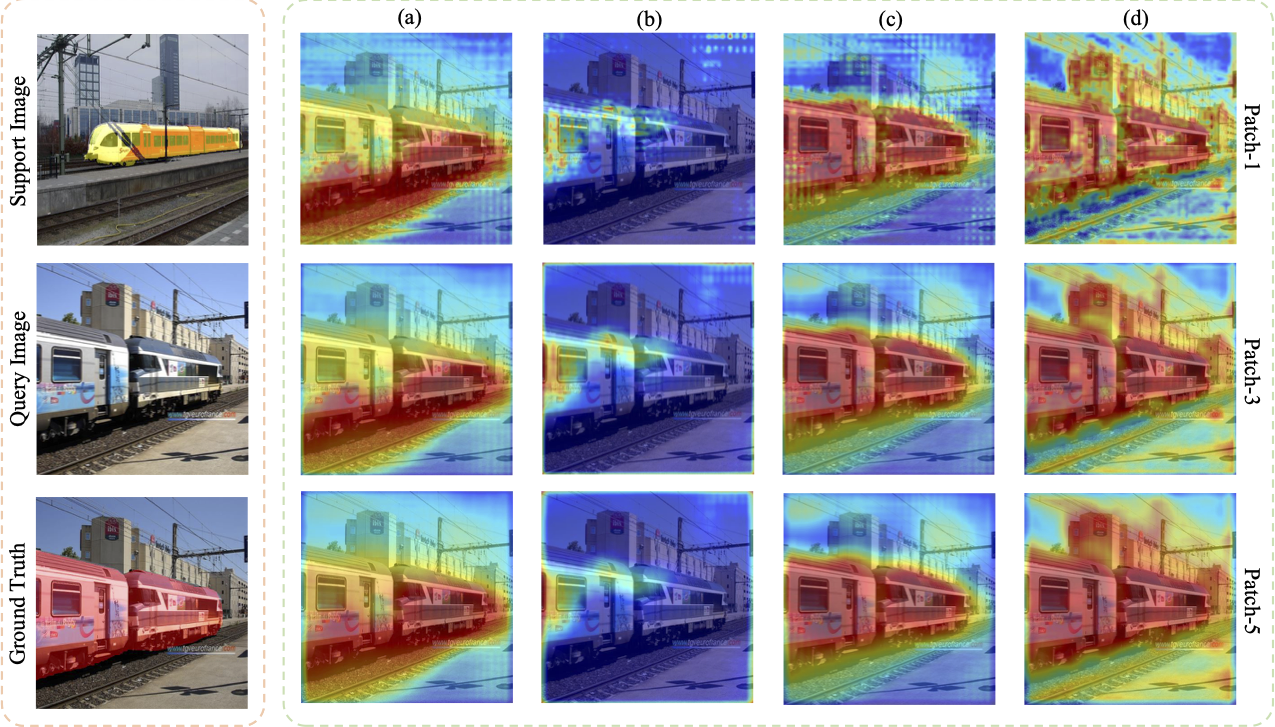}
    \end{minipage}       
    \caption{Visual comparison between prior masks generated by different sources and patch sizes: (a) original high-level features; (b) high-level features after projection; (c) high-level features after projection and NSM; (d) middle-level features after projection and NSM. Figures of (a) with patch 1 are the prior masks adopted by the original PFENet. Specifically, the prior masks of (a) maintain the basic generalization ability and can partially identify the unseen objects on the query images, while the linear channel compression deprecates the ones from (a) to (b) and images of (b) tend to erroneously highlight the background regions objects belonging to base classes used for training. Rectification is achieved by the proposed Noise Suppression Module (NSM) as demonstrated by figures of (c) that are generally more visually attractive than the two aforementioned sources. Also, introducing broader contextual information further enhances the prior masks, as manifested by the comparison between patch sizes 1, 3 and 5. Finally, the comparison between (c) and (d) shows that the prior masks yielded by middle-level features can supplement more local structural details but they are not as semantically precise as that of (c), hence PFENet++ incorporates both (c) and (d).}
    \label{fig:visual_compare_prior}
\end{figure*}

\subsection{Results}
\label{sec:compare_with_sota}
The quantitative results of three backbone networks are shown in Tables~\ref{tab:compare_sota_class},~\ref{tab:compare_sota_class_coco} and~\ref{tab:compare_sota_fss1000} where PFENet++ significantly outperforms all its competitors on both PASCAL-5$^i$, COCO-20$^i$ and FSS-1000, manifesting the superiority of the proposed CAPM and NSM. By adequately exploiting the context information during the prior mask generation process, PFENet++ even advances PFENet by 8-10 mIoU on 5-shot evaluations in terms of both class mean IoU and FB-IoU.  

Besides, it is noteworthy that PFENet++ achieves much better performance without remarkably sacrificing its efficiency since it can process 12.7 frames of size $473 \times 473$ per second with ResNet-50 backbone, while the original PFENet processes 14.8 frames per second on the same NVIDIA RTX 2080Ti GPU. As for the learnable parameters, the original PFENet has 10.8M while the proposed designs only bring 1.4M increase thus PFENet++ has 12.2M learnable parameters in total.  The qualitative results are shown in Figure~\ref{fig:visual_compare} where the results of PFENet++ are much better than that of the original PFENet. 

As for the impressive improvement from 1- to 5-shot settings, compared to PFENet, we conjecture that it can be attributed to the better use of high-level features, since more high-level contextual hints are leveraged by the proposed CAPM and NSM keeps the essence in a class-agnostic way for combating the over-fitting issues. Similar improvement can be observed in recent work RePRI where the high-level features are directly optimized by specific objectives designed for few-shot segmentation task. Thus we believe further exploiting the potentials of high-level features without impairing the generalization ability can be a promising future directions.

\subsection{Ablation Study}
\label{sec:ablation_study}
In this section, extensive experiments are presented to inspect the effectiveness of our two major contributions: Context-Aware Prior Mask (CAPM) and Noise Suppression Module (NSM). Experiments are conducted on PASCAL-5$^i$ with ResNet-50 backbone without specification. The different visual effects brought by the proposed CAPM and NSM are presented in Fig.~\ref{fig:visual_compare_prior}. All compared models in this section are re-trained for fair comparisons.

\begin{table}[!t]
    \caption{Comparison between different context enrichment modules.}
    \centering
    \tabcolsep=0.5cm
    {
        \begin{tabular}{ l |  c  c  }
            \toprule
            Methods 
            & 1-Shot & 5-Shot 
             \\
            \specialrule{0em}{0pt}{1pt}
            \hline
            \specialrule{0em}{1pt}{0pt}  
            Baseline (PFENet)  & 61.3 & 63.0 \\  
            + NSM  & 63.2 & 66.8   \\  
            + CAPM  & 58.6 & 59.7    \\  
            + NSM + PPM~\cite{pspnet}  & 51.3 & 53.2   \\   
            + NSM + ASPP~\cite{aspp}  & 55.3 & 56.5    \\  
			+ NSM + GAU~\cite{Zhang_2019_ICCV}  &59.6 & 61.5    \\              
            + NSM + CAPM  & \textbf{64.9} & \textbf{69.9}  \\  

            \bottomrule            
        \end{tabular}
    }
    \label{tab:ablation_study}
\end{table}

\begin{table}[!t]
	\caption{Ablation study on the effectiveness of NSM and CAPM. We note that CAPM = $\{1\}$ means there is no context-aware prior mask generation because only 1$\times$1 patch is adopted.  CAPM$=\{1,3,5\}$ means $|M|=3$ in Eq.~\eqref{eqn:multi_patch_similarity} and patch sizes are 1, 3 and 5. 
	}  
	\label{tab:scale_capm}       
	\centering
	\tabcolsep=0.4cm
	{
		\begin{tabular}{ 
				l |
				c
				c |
				c 
				c }
			\toprule
			Exp. &
			NSM & 
			CAPM & 
			1-Shot  &
			5-Shot   \\

			\specialrule{0em}{0pt}{1pt}
			\hline
			\specialrule{0em}{1pt}{0pt}
			
			\bignum{1} &
			-  & 
			\{1\} & 
			61.3 &           
			63.0 \\             
			
			\bignum{2} &
			-  & 
			\{1,3,5\} & 
			58.6 &           
			59.7  \\        
			
			\bignum{3} &
			\Checkmark  & 
			\{1\} & 
			63.2 &           
			66.8  \\                  
			
			\bignum{4} &
			\Checkmark  & 
			\{3\} & 
			62.9 &           
			66.1  \\                 
			
			\bignum{5} &
			\Checkmark  & 
			\{5\} & 
			62.9 &           
			66.2  \\           
			
			\bignum{6} &
			\Checkmark  & 
			\{1,3\} & 
			64.4 &           
			68.5 \\               
			
			\bignum{7} &
			\Checkmark  & 
			\{3,5\} & 
			63.9 &           
			68.0 \\                
			
			\bignum{8} &
			\Checkmark  & 
			\{1,3,5\} & 
			64.9 &           
			69.9   \\            
			
			\bignum{9} &
			\Checkmark  & 
			\{1,3,5,7\} & 
			64.8 &           
			69.4  \\

			\bottomrule                                   
		\end{tabular}
	}
\end{table}

\pamiparagraph{Noisy support features are the evils.}
NSM aims at alleviating the adverse effects brought by noisy support features during the prior mask generation process. Even the regional matching is not adopted in PFENet, the noisy support features still cause undesired high correlation responses with the query features on the prior mask. 
By applying NSM, it brings considerable improvement to the vanilla PFENet in Table~\ref{tab:ablation_study}.

\pamiparagraph{Context is essential but could also become evil.}
The original PFENet only leverages the element-to-element correspondence for yielding the prior mask while it overlooks the contextual information that could help better identify the region of interest. 
Therefore, as shown by Exp.~\bignum{3} \& \bignum{6} in Table~\ref{tab:scale_capm}, the result obtained from the patch-wise correlations (CAPM=$\{1, 3\}$) is better than the original 1 to 1 correlation (CAPM=$\{1\}$) adopted in PFENet because the former is more aware of the hints hiding in the surroundings. 

Additionally, since the large patch captures nearby context and the small patch mines finer details, the combination of them might be even better. The results of Exp.~\bignum{6}-\bignum{8} show that by mining extra cues from the nearby regions with different distances, the proposed multi-patch regional matching scheme of CAPM introduces further advancements to the models incorporated with single level matching schemes such as Exp.~\bignum{3}-\bignum{5}, but enlarging the patch size to 7 does not harvest additional improvement as shown in Exp.~\bignum{9}.

However, every rose has its thorn so does the contextual information since it might introduce unfiltered information that gives a rise to trivial responses on the prior masks. As demonstrated by the Exp.~\bignum{2}, if the NSM is not equipped for screening those noises induced by the additional contextual information, the performance is notably impaired, compared to Exp.~\bignum{1} and Exp.~\bignum{8}. With this finding, we can conclude that both CAPM and NSM are mutual-complementary and indispensable so as to achieve a promising performance.

\pamiparagraph{Comparison with other context enrichment modules.}
The purpose of CAPM is to leverage more contextual information during the prior generation process, thus CAPM performs regional matching with different patch sizes in Eq.~\eqref{eqn:multi_patch_similarity} to form correlation matrices that will be then processed by NSM for yielding the final prior mask.

Alternatively, directly applying context enrichment modules like Pyramid Pooling Module (PPM)~\cite{pspnet}, Atrous Spatial Pyramid Pooling (ASPP) module~\cite{aspp} and Graph Attention Unit (GAU)~\cite{Zhang_2019_ICCV} to the extracted high-level feature map seems to be another attainable way that provides a wider scope for individual elements in the one-to-one correlation calculation process as Eq.~\eqref{eqn:similarity_matrix}.

Experiments in Table~\ref{tab:ablation_study} show that, compared to the results of NSM+CAPM, both PPM and ASPP fall short of the generalization ability to unseen categories, because CAPM as well as the origin prior mask transform the high-level semantic information into the class-insensitive formats, \ie, prior masks whose values range from 0 to 1, to avoid the over-fitting issue. However, PPM and ASPP directly apply parameterized modules that are prone to overly rely on the high-level features to yield high-quality prior masks on base classes during training, leading to a rather inferior performance on novel classes.

Different from ASPP and PPM, the key component of PGNet~\cite{Zhang_2019_ICCV}, i.e., GAU,  aims at establishing regional correlations between query-support features specifically, so it applies several average pooling to the features followed by independent Graph Attention Units (GAU) to between the query and support individual elements at different levels. Though GAU is found beneficial in enhancing the middle-level features, two issues may inhibit the application to enhance the prior mask obtained from the high-level features: 1) Average pooling loses essential local information contained in the query features; 2) GAU directly projects the query and support features, thus it leads to over-fitting issues as ASPP and PPM.

Concretely, inherently different from the multi-scale schemes achieved by different average pooling sizes in GAU, the proposed CAPM models query-support correlations more meticulously by measuring the inter-patch similarities, keeping the local hints intact within the patches. Compared to the average pooling operations used by GAU for establishing the compressed regional correlations, patch-wise correlation adopted by CAPM retains local details better and thus it is more effective in revealing extra useful cues in different regions.
On the other hand, dense attention modules are directly applied to query and support features in GAU, so it is more likely to overfit the witnessed base classes. While, the proposed CAPM is with no learnable parameters and NSM only takes the class-agnostic statistics regarding the spatial responses as the input so as to ameliorate the over-fitting issue.

\begin{figure}
	\centering
	\begin{minipage}   {1.0\linewidth}
		\centering
		\includegraphics [width=1\linewidth] 
		{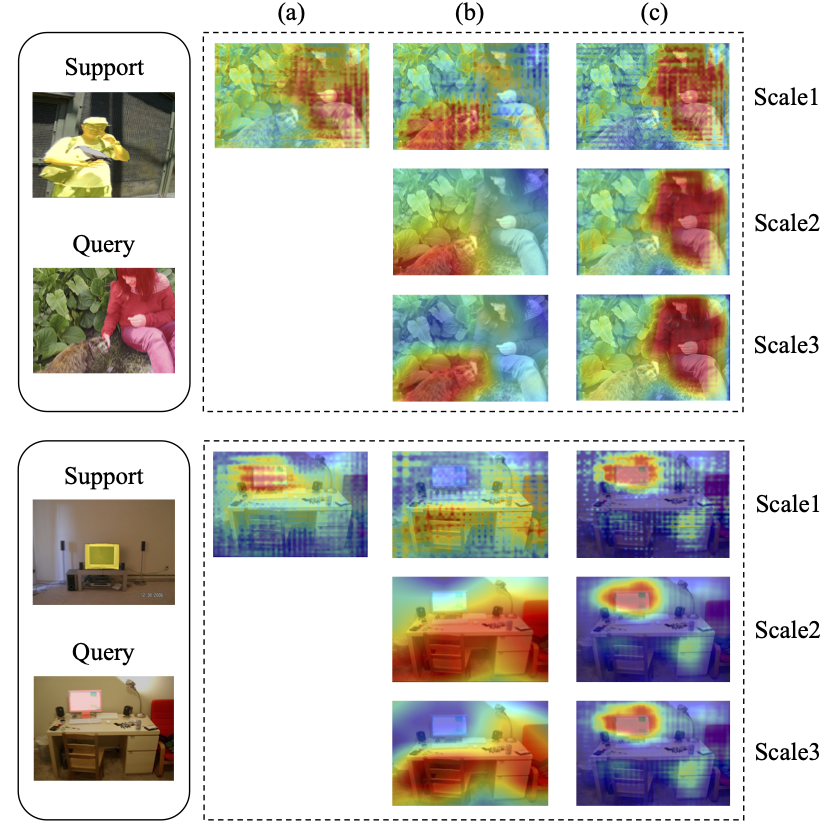}
	\end{minipage}
	\vspace{-0.2cm}    
	\caption{ 
		Qualitative comparisons of (a) the original PFENet's prior masks, (b) prior masks yielded with GAU, (c) the proposed context-aware prior mask.}
	\label{fig:pgnet}
\end{figure}

As the results shown in Table~\ref{tab:ablation_study} where we incorporate GAU to enhance the query features before yielding the prior mask, GAU is not helpful in enhancing the prior mask. Also, in Figure~\ref{fig:pgnet}, the visualizations yielded by (b) are more prone to high-light the objects belonging to the base classes in the background. 

\pamiparagraph{Middle-level features can be also leveraged for prior generation.}
The high-level features have more semantic cues while the middle-level ones encode more spatial details, thus the latter could be further leveraged to supplement the prior masks yielded by the high-level features. Due to the lack of an effective noise suppression mechanism, as shown in Exp.~\bignum{1} and Exp.~\bignum{2} in Table~\ref{tab:middle_and_high}, directly applying the prior masks yielded by the middle-level features is not beneficial but might potentially degrade the performance. Contrarily, by the virtue of the proposed NSM, the prior masks yielded by the middle-level features are also able to indicate the region of interest from a more local perspective, and the comparison between Exp.~\bignum{3} and Exp.~\bignum{4} in Table~\ref{tab:middle_and_high} manifests the improvement to the final performance. The visual comparison between the prior masks yielded by high- and middle-level features are shown in Fig.~\ref{fig:visual_compare_prior} (c)-(d).
Exp.~\bignum{5} proves the necessity of the high-level features by comparing it with Exp.~\bignum{3} and Exp.~\bignum{4}. Also, the results can tell that the considerable improvement of PFENet++ from 1 to 5-shot mainly comes from the better use of the high-level features by comparing Exps.~\bignum{3}-\bignum{5}. 

 \begin{table}[!t]
    \caption{Results of applying CAPM on the middle- and high-level features with and without the proposed NSM. 
    }  
    \label{tab:middle_and_high}       
    \centering
    \tabcolsep=0.3cm
    {
        \begin{tabular}{ 
          l |
          c 
          c
          c |
          c 
          c }
            \toprule
            Exp. &
            NSM & 
            High & 
            Middle & 
            1-Shot  &
            5-Shot  \\

            \specialrule{0em}{0pt}{1pt}
            \hline
            \specialrule{0em}{1pt}{0pt}
        
            \bignum{1} &
            -  & 
            \Checkmark & 
            - & 
            61.3 &           
            63.0 \\              

            \bignum{2} &
            -  & 
            \Checkmark & 
            \Checkmark & 
            61.0 &           
            63.1 \\

            \bignum{3} &
            \Checkmark  & 
            \Checkmark & 
            - & 
            64.1 &        
            68.6  \\              

            \bignum{4} &
            \Checkmark  & 
            \Checkmark & 
            \Checkmark & 
            64.9 &           
            69.9\\

            \bignum{5} &
			\Checkmark  & 
			- & 
			\Checkmark & 
			60.2 &           
			60.7 \\                  
            \bottomrule                                   
        \end{tabular}
    }
\end{table}

\pamiparagraph{Design options for higher efficiency. }
The multi-patch regional matching involves extra computation, causing lower efficiency. Table~\ref{tab:efficient_choice} shows the effects of different structural options for achieving an efficient design, in terms of performance, speed and the number of learnable parameters.

With an eye towards a faster implementation, first, we prefer a slim channel number because high-dimensional features slow the correlation calculation. As shown in Exp.~\bignum{1}-\bignum{2}, by applying a simple 1$\times$1 convolution to reduce both the query and support high-level feature maps' dimensions from 2048 to 256 before performing the regional matching in Eq.~\eqref{eqn:patch_similarity}, the inference becomes considerably faster (from 3.8 to 9.1 FPS). Additionally, the compressed features retain essential information without deprecating the performance. 

Also, we find that applying a 2$\times$2 average pooling on the support feature maps further accelerates the regional matching process in Eq.~\eqref{eqn:patch_similarity} without sacrificing the performance as verified by Exp.~\bignum{2}-\bignum{3} in Table~\ref{tab:efficient_choice}. We note that the decrease in Params is caused by the parameterized module $\Theta$ in NSM whose input and output dimensions are determined by the spatial size of support feature map.
However, compared to Exp.~\bignum{3}, Exp.~\bignum{4} indicates that spatially compressing the query feature maps yields inferior results because it leads to direct information loss on the target query feature maps.

\pamiparagraph{The benefits of down-sampling. }
Down-sampling has two benefits. Firstly, it is used to reduce the computational cost introduced by the multi-scale patch matching, and the effects have been shown in Table 7 of the paper. On the other hand, it can introduce broader contextual information for enhancing the prior mask. To verify this claim, we conduct another experiment without down-sampling but with enlarged patch sizes $\{1, 5, 9\}$ in Table~\ref{tab:efficient_choice}. It can be observed that, compared to the original scheme adopted in the paper with patches $\{1, 3, 5\}$, the model implemented with $\{1, 5, 9\}$ has considerably lower inference speed (12.7 \textit{v.s.} 7.9). However, similar to PFENet++ with down-sampling,  the model incorporating larger patch sizes $\{1, 5, 9\}$ still outperforms the PFENet++ without down-sampling the support features, \textit{i.e.}, PFENet++(Full), because additional contextual information can be introduced by either down-sampling or larger patch sizes, yielding better performance. As for the reason why $\{1, 5, 9\}$ is inferior to the PFENet++ with down-sampling and $\{1, 3, 5\}$, we think explicit larger patch matching may bring extra noises, causing additional difficulties for NSM.

\pamiparagraph{The effects of the rectification module $\Theta$. }
DeepEMD~\cite{deepemd} adopts dot-product between each support feature and all query features in order to high-light the importance of the individual support element, which is analogous to the operation in the first step of NSM. The difference is that DeepEMD directly takes the average of the responses with a normalization operation as the weights for different support elements, while PFENet++ adopts a learnable module $\Theta$ for dynamic rectification based on the global responses. Though it is found rather effective in few-shot classification, without the rectification module, merely adopting the operation of DeepEMD in dense semantic segmentation is sensitive and vulnerable to noisy samples, as demonstrated in Figure~\ref{fig:statistic_compare}. Also, we have also tried by replacing the NSM with the operation of DeepEMD, and the model obtains 58.6 and 61.4 mIoU in 1- and 5-shot settings respectively. 

\pamiparagraph{The necessity of the position-sensitive operations in $\Theta$.}

We consider a key aspect that \textit{the spatial information still exists in $\mathcal{R}_{\psi}\in \mathbb{R}^{1\times h_s w_s}$}, since the compression performed by the concentrator $\Psi$ does not result in the loss of spatial information within $\mathcal{R}_{\psi}$ as it is obtained by flattening $h_s \times w_s$ to $1\times h_s w_s$. This preservation of positional and spatial information can be likened to the transformation performed by the vision transformer, where each image is converted into $16\times16$ $d$-dimensional tokens, specifically, flattened into the vector whose shape is $d\times 196$. 
In contrast to the vision transformer's utilization of position embeddings for conveying positional information, NSM employs the position-sensitive linear layers as the rectification module $\Theta$ to analyze the obtained correlation $\mathcal{R}_{\psi}\in \mathbb{R}^{1\times h_s w_s}$ from the concentrator $\Psi$. Therefore, it is crucial to note that \textit{without the position-sensitive $\Theta$, the effective filtering out of potential noises present in $\mathcal{R}_{\psi}$ would not be achievable}. 

To illustrate this, consider a scenario where a specific location in $\mathcal{R}_{\psi}$ exhibits high activation, while its surrounding area displays low activation. In such a case, this location is likely to be a noisy element that NSM aims to suppress in the support sample. However, without the adoption of the position-sensitive module $\Theta$, we lack the means to analyze each location with the context of its surroundings or broader area, thereby impeding our ability to determine whether it constitutes noise or not. 

To support our claim, we have conducted several experiments, and the results are shown in Table~\ref{tab:exp_nsm}.
Specifically, the position-agnostic operations of self-attention (`Self-Attn') and pooling (`Pooling') demonstrate clear lower performance compared to the proposed NSM. This discrepancy can be attributed to their limited capability in effectively utilizing spatial information to identify potential noisy samples within $\mathcal{R}_{\psi}$, as discussed earlier. When incorporating the positional embedding to the self-attention module, as shown by the results of `Self-Attn + Pos', it shows better results than the aforementioned position-agnostic counterparts, but it is still inferior to that of NSM.
Furthermore, as in `Self-Attn + NSM', we investigate the effects of adding self-attention to enhance the interaction between different elements in $\mathcal{R}$ prior to passing it to NSM. However, it does not yield improved performance. Thse findings emphasize the crucial role played by the \textit{position-sensitive} linear layers employed in NSM.

\begin{table}
  \caption{The results of different methods to obtain $\mathcal{R}_{\psi}$. 
    `Pooling' indicates that NSM does not employ a linear layer but instead directly applies pooling to $\mathcal{R}$. `Self-Attn' refers to the utilization of a self-attention layer to process $\mathcal{R}$, which is treated as $h_{s}w_{s}$ vectors with individual shapes $[1 \times h_{q}w_{q}]$. The final $\mathcal{R}_{\psi}$ of `Self-Attn' is obtained through the application of the concentrator $\Psi$, serving as the NSM. Additionally, `Self-Attn + Pos' adopts positional embeddings for individual elements, and `Self-Attn + NSM' incorporates the self-attention layer before NSM.  }
    \centering
    \tabcolsep=0.7cm
    \begin{tabular}{l|c|c}
    \hline
        Methods & 1-Shot & 5-Shot\\
        \hline
        NSM & 64.9 & 69.9 \\
        Pooling & 58.6 & 59.7 \\
        Self-Attn & 59.9 &62.7 \\
        Self-Attn + Pos & 63.6 & 66.1 \\
        Self-Attn + NSM & 64.8 & 69.2\\
    \hline
    \end{tabular}
    \label{tab:exp_nsm}
\end{table}

\pamiparagraph{More about the prior mask. }
The prior mask is used as an input of FEM to provide prior knowledge of the existence of the target class, so the quality is essential for the final performance. If we directly use the ground-truth mask to replace the estimated prior mask, the 1- and 5-shot results are directly boosted to 98.0 and 99.0 mIoU respectively, manifesting the importance of the prior mask. Besides, we have also examined another model that directly outputs the prior masks as the final prediction, and we surprisingly find that the prior mask alone can even achieve 59.3 and 62.4 mIoU in 1 and 5-shot settings without the help of the decoder with FEM. 

 \begin{table}[!t]
    \caption{Ablation study on the proposed Noise Suppression Module (NSM). `Ch' represents the channel compression. `Sp-S' and `Sp-Q' are spatial down-sampling applied on support and query features respectively. `FPS' denotes the number of frames processed per second. `Params' is the number of learnable parameters. Exp.~\bignum{2}$^*$ means larger patches are adopted without down-sampling ($\{1,5,9\}$ instead of $\{1,3,5\}$) for leveraging broader contextual information.
    }  
    \label{tab:efficient_choice}       
    \centering
    \tabcolsep=0.2cm
    {
        \begin{tabular}{ 
          l |
          c
          c c |
          c 
          c | c c }
            \toprule
            Exp. &
            Ch & 
            Sp-S & Sp-Q & 
            1-Shot  &
            5-Shot & FPS & Params (M) \\

            \specialrule{0em}{0pt}{1pt}
            \hline
            \specialrule{0em}{1pt}{0pt}
            
            \bignum{1} &
            - & 
            - & 
            - & 
            62.2 &  
            65.0 & 
            3.8 & 13.1 \\     
            
            \bignum{2} &
            \Checkmark & 
            - & 
            - & 
            63.1 &  
            66.7 & 
            9.1 & 13.6 \\     
         
            \bignum{2}$^*$ &
            \Checkmark & 
            - & 
            - & 
            64.3 &  
            68.0 & 
            7.9 & 13.6 \\            
            
            \bignum{3} &
            \Checkmark & 
            \Checkmark & 
            - & 
            \textbf{64.9} &           
            \textbf{69.9} & 
            12.7 & 12.2 \\              
            
            \bignum{4} &
            \Checkmark & 
            \Checkmark & 
            \Checkmark & 
            64.2 &           
            68.9 & 
            14.1 & 12.2 \\

            \bottomrule                                   
        \end{tabular}
    }
\end{table}

\subsection{Extensions}
\pamiparagraph{Generalization on totally unseen categories. }
The majority of unseen categories for the evaluations on PASCAL-5$^i$ and COCO-20$^i$ has been included in the class set of ImageNet~\cite{imagenet}, therefore the ImageNet pre-trained backbones have already witnessed these novel classes, even if their corresponding samples and pixel-wise annotations are not available during the model training. 
To further demonstrate the effectiveness of the proposed CAPM and NSM, the comparison of the classes that are NOT contained in the ImageNet is more convincing. 

In PASCAL-5$^i$, the category `person' is not included in ImageNet. Thus the original prior mask adopted in PFENet might fail to locate the region belonging to `person' since the context is not well exploited. By incorporating the proposed CAPM and NSM, Ours (48.56) significantly outperforms the prior mask used in the original PFENet (15.81) in terms of mIoU on `person'.

However, a single class is insufficient to show the generalization ability. FSS-1000~\cite{fss1000} is a benchmark that sets up 1,000 classes for few-shot segmentation among which we select 420 classes \footnote{The paper of FSS-1000 wrote that there are 486 classes not belonging to any existing datasets. The author of FSS-1000 has clarified in an email that they ``have made incremental changes
to the dataset to improve class balance and label quality so the number
may have changed. Please do experiments according to the current
version.”} 
 from FSS-1000 as the novel class set for testing since they are not in the class set of ImageNet and the rest classes are used as training class set.
As shown in Table~\ref{tab:unseen_fss1000}, the proposed CAPM and NSM still demonstrate their efficacy by well generalizing to the 420 unseen categories that are even not included in the large scale dataset for pre-training the backbone.

\pamiparagraph{Applications on other frameworks.}
The enhanced prior mask, as a model-agnostic pixel-wise indicator that has no special structural constraints, should be able to bring decent performance gain to different frameworks, not only to PFENet. To investigate the generalization ability of the proposed CAPM and NSM, we additionally apply them on two recent state-of-the-art methods SCL~\cite{scl} and ASGNet~\cite{asgnet}. 

Specifically, SCL proposes a self-guided learning approach that creates main and auxiliary support vectors to facilitate extracting the discriminative information. Differently, ASGNet, by applying superpixel-guided clustering (SGC) and guided prototype allocation (GPA), alleviates the ambiguities caused by using one prototype to represent all the information. In ASGNet, more than one representative prototypes are aggregated and then selected as the final support vectors for feature matching. 

Both two methods do not well exploit the prior knowledge hidden in high-level features, thus the proposed techniques significantly boost their performance as shown in Table~\ref{tab:application_to_other_models}. The implementation is similar to that of PFENet++, we simply attach a prior generation branch with the proposed CAPM and NSM, and then concatenate the new prior masks to the features before making the final predictions. Our implementations of ASGNet and SCL will be also made publicly available along with that of PFENet++.

\begin{table}
    \caption{Foreground IoU results on totally unseen classes of FSS-1000~\cite{fss1000}. PM denotes the original prior mask generation method used in PFENet~\cite{pfenet}. }
    \centering
    \tabcolsep=0.5cm
    {
        \begin{tabular}{ l |  c  c  }
            \toprule
            Methods
            & 1-Shot & 5-Shot \\
            \specialrule{0em}{0pt}{1pt}
            \hline
            \specialrule{0em}{1pt}{0pt}
            Baseline & 79.7 & 80.1 \\  
            PM~\cite{pfenet} & 80.8 & 81.4 \\  
            Ours  & \textbf{83.6} & \textbf{84.8} \\ 
            \bottomrule            
        \end{tabular}
    }
    \label{tab:unseen_fss1000}
\end{table}

\begin{table}
    \caption{Results of applying CAPM and NSM to  ASGNet~\cite{asgnet} and SCL~\cite{scl} on PASCAL-5$^i$. We note that SCL has already incorporated the vanilla prior mask proposed by PFENet, so only ASGNet (Vanilla Prior) is presented.}
    \centering
    \tabcolsep=0.5cm
    {
        \begin{tabular}{ l |  c  c  }
            \toprule
            Methods
            & 1-Shot & 5-Shot \\
            \specialrule{0em}{0pt}{1pt}
            \hline
            \specialrule{0em}{1pt}{0pt}
            ASGNet (original)  & 59.3 & 63.9 \\  
            ASGNet (reproduced) & 59.5 & 64.2 \\ 
            ASGNet (Vanilla Prior) & 60.3 &64.7 \\              
            ASGNet + ours & \textbf{63.4} & 67.4 \\  
            
            \specialrule{0em}{0pt}{1pt}
            \hline
            \specialrule{0em}{1pt}{0pt}  
            
            SCL (original)  & 61.8 & 62.9 \\  
            SCL (reproduced) & 61.3 & 62.6 \\  
            SCL + ours & \textbf{64.5} & \textbf{66.9} \\              
            \bottomrule            
        \end{tabular}
    }
    \label{tab:application_to_other_models}
\end{table}

\section{Conclusion}
\label{sec:conclu}
In this paper, we present a novel framework, named PFENet++, to tackle the few-shot semantic segmentation problem. Different from PFENet that only considers the maximum correspondence values between individual query and support features, PFENet++ incorporates broader contextual information to yield the Context-aware Prior Mask (CAPM) and adopts a lightweight Noise Suppression Module (NSM) that further improves the generalization ability on unseen classes by effectively selecting representative support features to alleviate the adverse effects brought by noisy responses. Not only surpassing the original PFENet without compromising much efficiency, but PFENet++ also significantly outperforms previous few-shot segmentation methods and
achieves new state-of-the-art performance on both PASCAL-$5^i$, COCO-$20^i$ and FSS-1000 benchmarks. Finally, extensive experiments are conducted to investigate the contributions of individual components of the proposed frameworks, and they are also conducive to different frameworks. We hope that PFENet++ can be a strong baseline approach to few-shot segmentation, and inspires future work such that the full potential of prior mask guidance can be further exploited.

{\small
		\bibliographystyle{ieee_fullname}
		\bibliography{egbib}
}

\end{document}